\pdfoutput=1

\documentclass[11pt]{article}

\usepackage[preprint]{acl}

\usepackage{times}
\usepackage{latexsym}

\usepackage[T1]{fontenc}

\usepackage[utf8]{inputenc}

\usepackage{microtype}

\usepackage{inconsolata}

\usepackage{graphicx}
\usepackage{booktabs}
\usepackage{lipsum}
\usepackage{covington}
\usepackage{CJKutf8}
\usepackage{hwemoji}
\usepackage{epigraph}
\title{Not quite Sherlock Holmes: Language model predictions do not reliably differentiate impossible from improbable events}

\author{
 \textbf{James A. Michaelov\textsuperscript{1,2,3}},
 \textbf{Reeka Estacio\textsuperscript{3}},
 \textbf{Zhien Zhang\textsuperscript{3}},
 \textbf{Benjamin K. Bergen\textsuperscript{3}}
\\
 \textsuperscript{1} Department of Brain and Cognitive Sciences, MIT  ~~~ \textsuperscript{2} MIT Libraries CREOS\\
 \textsuperscript{3} Deparmtent of Cognitive Science, UCSD
\\
  \texttt{jamic@mit.edu}, ~~ \texttt{\{rdestaci,zhz067,bkbergen\}@ucsd.edu} \\}

\begin{document}
\maketitle
\begin{abstract}
Can language models reliably predict that possible events are more likely than merely improbable ones? By teasing apart possibility, typicality, and contextual relatedness, we show that despite the results of previous work, language models' ability to do this is far from robust. In fact, under certain conditions, all models tested---including Llama 3, Gemma 2, and Mistral NeMo---perform at worse-than-chance level, assigning higher probabilities to impossible sentences such as `the car was given a parking ticket by the brake' than to merely unlikely sentences such as `the car was given a parking ticket by the explorer'.

\includegraphics[width=1em,height=1em]{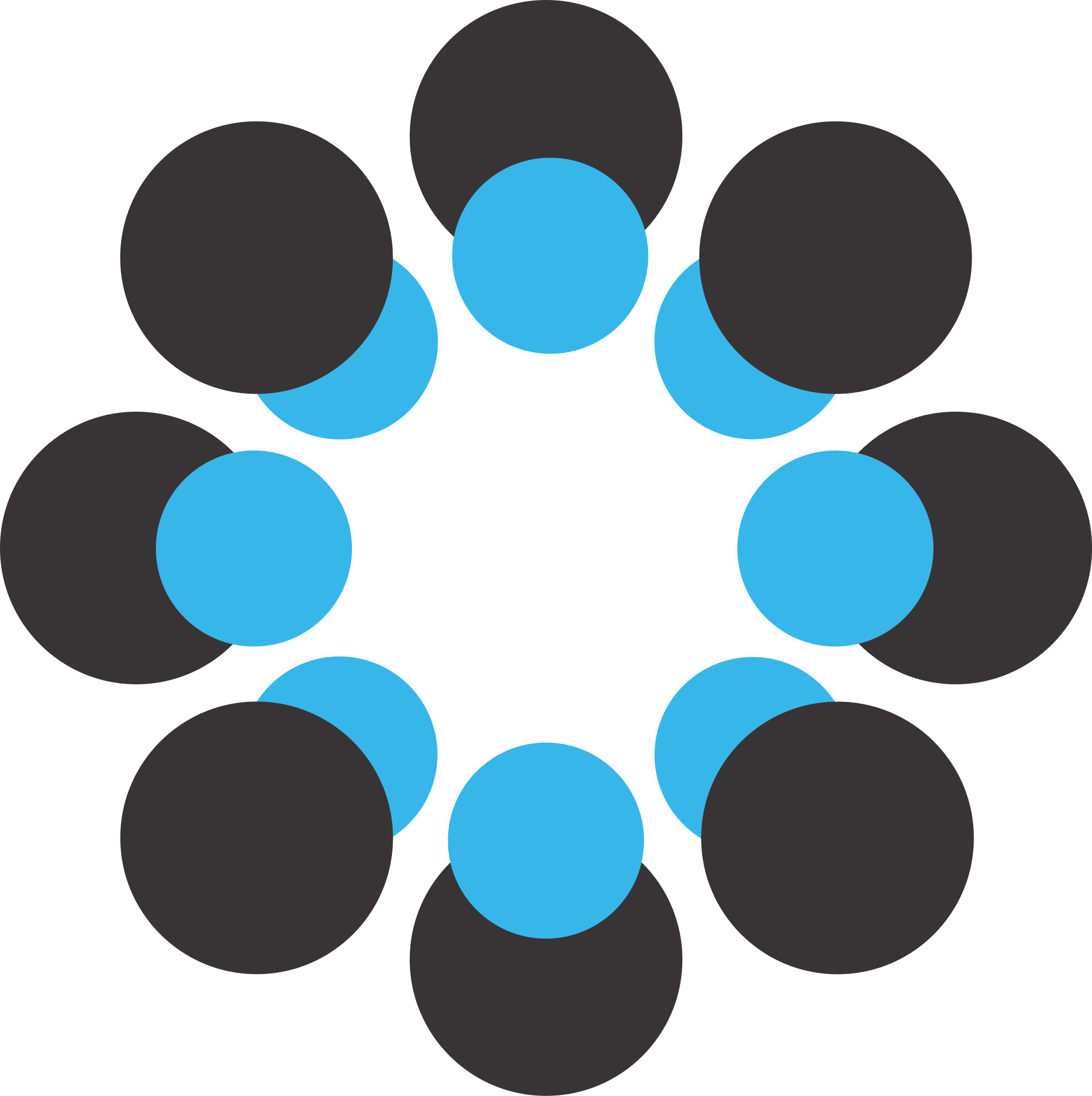}\hspace{.15em}
\parbox{\dimexpr\linewidth-7\fboxsep-7\fboxrule}{\href{https://osf.io/r6xns/?view_only=0567164a44f64530bde24c3bc5f1ddbd}{Data, Code, and Analyses}}

\end{abstract}

\section{Introduction}

\epigraph{\normalsize\itshape``How often have I said to you that when you have  eliminated the impossible, whatever remains, \textit{however improbable}, must be the truth?''\vspace{0.5em}}{\normalsize Sherlock Holmes, \textit{The Sign of the Four} \\\citep{doyle_1890_SignFour}}

\noindent Consider the following scenario:
\begin{subexamples}[preamble={Marissa forgot to bring her pillow on her camping trip. As a substitute for her pillow, she filled up an old sweater with...},postamble={\citep{glenberg_2000_SymbolGroundingMeaning}}]
\label{ex:glenberg}
\item clothes
\item leaves
\item water
\end{subexamples}
\textit{Clothes} is the obvious best answer here, and that most preferred by humans when presented with the options shown above (see \citealp{glenberg_2000_SymbolGroundingMeaning}). Nonetheless, \textit{leaves} is also a possible continuation. In fact, \citet{glenberg_2000_SymbolGroundingMeaning} find that experimental participants rate \textit{leaves} to be significantly more sensible than \textit{water}, despite the fact that the sentence \textit{as a substitute for her pillow, she filled up an old sweater with leaves} describes a highly unusual event.

This example (and indeed, all of \citealp{glenberg_2000_SymbolGroundingMeaning}) hinges specifically on the physical properties of objects and their `affordances' (i.e., what they can be used to do; \citealp{gibson_1966_SensesConsideredPerceptual,gibson_1979_EcologicalApproachVisual}). But it exemplifies a broader behavior that humans exhibit largely unconsciously, namely, using knowledge of the world to distinguish between the impossible and the merely atypical. In this study, we focus on whether language models also have the capability do so.

A growing body of work on world knowledge and event understanding investigates whether language models can select the most likely or plausible of a set of possibilities (e.g., \citealp{zellers_2019_HellaSwagCanMachine,sakaguchi_2020_WinoGrandeAdversarialWinograd,bisk_2020_PIQAReasoningPhysical}; see \autoref{ssec:typicality}). But this work often conflates different ways that the `incorrect' options can be incorrect; some are impossible, but others are simply less typical. 
While finding that models are better or worse at identifying likely events is in itself important, it is only one component of robust event understanding. In fact, any use of language models in practice, whether as part of larger systems or directly as chat assistants, is likely to involve novel and unexpected situations. Thus being able to tell the difference between the impossible and the merely unlikely is crucial, especially in critical domains where the use of language models has been suggested, such as medicine \citep[see, e.g.][]{vanveen_2024_AdaptedLargeLanguage,lievin_2024_CanLargeLanguage,singhal_2025_ExpertlevelMedicalQuestion}.

For this reason, we argue that it is important to also assess the extent to which language models are able to differentiate between possible but improbable events and impossible ones. We call this the \textbf{Sherlock Holmes Task} after the famous epigraphic quote \citep{doyle_1890_SignFour}.

In this study, we evaluate language models on a specific version of the Sherlock Holmes task, asking whether they are able to reliably assign a higher probability to sentences describing possible (but in some cases, atypical or unlikely) events (e.g., \textit{the car was given a parking ticket by the \textbf{delinquent}}; \citealp{vega-mendoza_2021_ConcurrentUseAnimacy}) than their impossible equivalents (e.g., \textit{the car was given a parking ticket by the \textbf{stamp}}). We 
 draw on experimental stimuli from previous research investigating how humans process such sentences \citep{vega-mendoza_2021_ConcurrentUseAnimacy,chow_2013_NoSemanticIllusions}. Prior work shows that language models assign impossible sentences with words semantically related to their context higher probabilities than equivalent sentences with unrelated words \citep{michaelov_2022_CollateralFacilitationHumans}. Thus, we also include a further adversarial component, specifically considering cases where impossible sentences include related words (e.g., \textit{the car was given a parking ticket by the \textbf{brake}}) and possible sentences include unrelated words (e.g., \textit{the car was given a parking ticket by the \textbf{explorer}}).

In contrast to previous work suggesting that language models have relatively good world knowledge and event understanding capabilities \citep[e.g.,][]{kauf_2023_EventKnowledgeLarge}, we find that both event atypicality and semantic relatedness lead to significant drops in performance. In the most highly adversarial case, where we compare the probabilities assigned to possible but atypical sentences with unrelated words to those assigned to impossible sentences with related words (e.g., \textit{the car was given a parking ticket by the \textbf{explorer}} vs. \textit{\textbf{brake}}), we find that all language models tested perform at or below chance---that is, they assign the impossible sentences a higher probability half or more of the time. We further find that this effect does not disappear with scale---in fact, smaller models often do better than their larger counterparts. Together, these results suggest that language models may rely on typicality and semantic relatedness cues when making predictions rather than actual world knowledge, and thus, that any previously-ascribed event understanding capabilities are far from robust.

\section{Background}
\subsection{Event Typicality}
\label{ssec:typicality}
Texts vary in the extent to which they describe a state of affairs that is likely and congruent with reality. In studies of human language comprehension, this has been operationalized in a variety of ways, including \textit{plausibility} \citep[e.g.,][]{paczynski_2012_MultipleInfluencesSemantic,vega-mendoza_2021_ConcurrentUseAnimacy}, sensibility \citep[e.g.,][]{glenberg_2000_SymbolGroundingMeaning}, \textit{typicality} \citep[e.g.,][]{urbach_2010_QuantifiersMoreLess}, \textit{combinability} \citep[e.g.,][]{chow_2013_NoSemanticIllusions}, and \textit{canonicality} \citep[e.g.,][]{nieuwland_2006_WhenPeanutsFall}. For consistency, in this paper we use the term \textit{event typicality} as a general term encompassing the general unifying idea behind these---the extent to which an event described is likely to occur.

One common way to evaluate language models' sensitivity to event typicality is in commonsense and physical reasoning benchmarks (e.g., \citealp{levesque_2012_WinogradSchemaChallenge,sakaguchi_2020_WinoGrandeAdversarialWinograd,talmor_2019_CommonsenseQAQuestionAnswering,bisk_2020_PIQAReasoningPhysical,zellers_2018_SWAGLargeScaleAdversarial,zellers_2019_HellaSwagCanMachine}), which often at least implicitly target this capability \citep{davis_2015_CommonsenseReasoningCommonsense,storks_2020_RecentAdvancesNatural}. Take, for example, the following item from the widely-used HellaSwag dataset \citep{zellers_2019_HellaSwagCanMachine}:

\begin{subexamples}[preamble={A woman is outside with a bucket and a dog. The dog is running around trying to avoid a bath. She...}]
\label{ex:hellaswag}
\item rinses the bucket off with soap and blow dry [\textit{sic}] the dog’s head.
\item uses a hose to keep it from getting soapy.
\item gets the dog wet, then it runs away again.
\item gets into a bath tub with the dog.
\end{subexamples}

In this example, continuation (c) is `correct'---it corresponds to the most typical sequence of events. And indeed, continuation (c) is the one that actually occurs in the ActivityNet Captions dataset \citep{krishna_2017_DenseCaptioningEventsVideos} from which the example is derived. However, the `incorrect' continuations, while less plausible and appropriate, also all describe possible continuations of the event in question---we don't see any truly impossible continuations such as \textit{[she] and the dog are chased by the bucket}.

Several studies have explicitly investigated event possibility \citep{michaelov_2023_CanPeanutsFall,kauf_2023_EventKnowledgeLarge,hanna_2023_WhenLanguageModels}. These studies generally base their analyses on experimental stimuli derived from human psycholinguistic studies where impossible sentences always involve animacy violations such as \textit{the peanut was in love} (\citealp{michaelov_2023_CanPeanutsFall,hanna_2023_WhenLanguageModels}; original stimulus from \citealp{nieuwland_2006_WhenPeanutsFall}), or \textit{the laptop bought the teacher} (\citealp{kauf_2023_EventKnowledgeLarge}; original stimulus from \citealp{fedorenko_2020_LackSelectivitySyntax}). 

These studies investigate whether language models can differentiate between possible and impossible events. Yet in all cases, the best continuation is a typical one. This is true even in the case of \citet{kauf_2023_EventKnowledgeLarge}, who investigate whether there is a difference between the performance of language models at distinguishing plausible from impossible events compared to distinguishing plausible from merely implausible (i.e., atypical) events. Thus, none of these studies are able to successfully tease apart the effects of possibility and typicality, which is important because it is precisely in atypical situations that it is important to distinguish between possible and impossible events.

\subsection{Semantic Relatedness}
A number of studies have shown that that the word predictions of both humans \citep[e.g.,][]{ettinger_2016_ModelingN400Amplitude,uchida_2021_ModelOnlineTemporalSpatial} and language models \citep[e.g.,][]{misra_2020_ExploringBERTsSensitivity,michaelov_2024_StrongPredictionLanguage} correlate with the word's semantic relatedness to its context. For example, \citet{misra_2020_ExploringBERTsSensitivity} find that adding words like \textit{airplane} (either alone or in a sentence context) before sentences like \textit{I wanted to become a...} increases the probability of BERT predicting words like \textit{pilot} compared to when adding a control word such as \textit{table}.

This tendency is in principle adaptive---it would be surprising if within an utterance or connected discourse, one were to encounter a sentence completely unrelated to anything mentioned previously \citep[see, e.g.,][]{grice_1975_LogicConversation,grice_1989_StudiesWayWords,sperber_1986_RelevanceCommunicationCognition}. Thus, contextual relatedness may often function as a reliable heuristic of what is to come next.

However, the evidence suggests that using contextual semantic relatedness as a partial basis for prediction may not always be beneficial. Consider, for example, the following text from a study by \citet{metusalem_2012_GeneralizedEventKnowledge}:

\begin{subexamples}[preamble={We're lucky to live in a town with such a great art museum. Last week I went to see a special exhibit. I finally got in after waiting in a long...}]
\label{ex:metusalem}
\item line
\item painting
\item toothbrush
\end{subexamples}

\citet{metusalem_2012_GeneralizedEventKnowledge} find that while \textit{painting} and \textit{toothbrush} are both impossible sentence continuations, and are equally unlikely to be offered up as possible continuations by experimental participants, \textit{painting} is more strongly predicted during the process of language comprehension, which \citet{metusalem_2012_GeneralizedEventKnowledge} argue is due to its relation to the event discussed in context (i.e., visiting an art museum). \citet{michaelov_2022_CollateralFacilitationHumans} replicate this finding in language models, finding that they also predict \textit{painting} to be more likely. While in this context, such a result may not be particularly harmful, as it amounts to a difference between two impossible continuations, it does suggest that reliance on contextual semantic relatedness could be problematic in other cases.

\subsection{Possibility}
We are only aware of two studies that directly test whether language models are able to successfully assign sentences describing possible but atypical events a higher probability than impossible ones. \citet{kauf_2023_EventKnowledgeLarge} look across a large set of sentences of comparable length and structure, finding that overall, impossible sentences are assigned lower probabilities than merely implausible ones. 

\citet{jones_2022_DistrubutionalSemanticsStill}, on the other hand, carry out a study on the full set of stimuli from \citet{glenberg_2000_SymbolGroundingMeaning}, as exemplified in (\ref{ex:glenberg}). Thus they directly investigate cases where the texts differ only by one word---a \textit{critical word}---that determines whether they are likely (as in \textit{she filled up an old sweater with \textbf{clothes}}; see (\ref{ex:glenberg})), possible but atypical (i.e., afforded; e.g., \textit{she filled up an old sweater with \textbf{leaves}}), or impossible (i.e., nonafforded; e.g., \textit{she filled up an old sweater with \textbf{water}}). \citet{jones_2022_DistrubutionalSemanticsStill} find that BERT and RoBERTa are not able to consistently assign the possible but atypical events a higher probability than the impossible ones, but (on average) GPT-3 does; though not to the extent that it can fully account for human sensibility judgments. 

\begin{CJK*}{UTF8}{gbsn}
\begin{table*}[t]
\small
\centering
\begin{tabular}{@{}llllll@{}}
\toprule
\textbf{Lang.} &
  \textbf{Sentence} &
  \textbf{Possibility} &
  \textbf{Typicality} &
  \textbf{Relatedness} & \\ \midrule
Eng. & The cure for the disease was discovered by the \underline{doctor}.  & ✅ \textbf{P}ossible & ✅ \textbf{T}ypical & ✅ \textbf{R}elated & (PTR)\\
        & The cure for the disease was discovered by the \underline{patient}.   & ✅ \textbf{P}ossible & ❌ \textbf{A}typical & ✅ \textbf{R}elated & (PAR)\\
        &   The cure for the disease was discovered by the \underline{guest}.    & ✅ \textbf{P}ossible & ❌ \textbf{A}typical & ❌ \textbf{U}nrelated & (PAU)\\
        &   The cure for the disease was discovered by the \underline{medication}.     & ❌ \textbf{I}mpossible & ❌ \textbf{A}typical  & ✅ \textbf{R}elated & (IAR)\\
        &   The cure for the disease was discovered by the \underline{stamp}.     & ❌ \textbf{I}mpossible & ❌ \textbf{A}typical  & ❌ \textbf{U}nrelated & (IAU)\\ \midrule
Mand. &
  
  \begin{tabular}[c]{@{}l@{}}高材生把数学题\underline{解答}了\\ `The student \underline{solved} the math problem'\end{tabular} &
  ✅ \textbf{P}ossible &
  ✅ \textbf{T}ypical &
  ✅ \textbf{R}elated& (PTR) \\
 &
  
  \begin{tabular}[c]{@{}l@{}}高材生把数学题\underline{挂起}了\\ `The student \underline{hung} the math problem'\end{tabular} &
  ✅ \textbf{P}ossible &
  ❌ \textbf{A}typical  &
  ❌ \textbf{U}nrelated& (PAU) \\
 &
  
  \begin{tabular}[c]{@{}l@{}}高材生把数学题\underline{难倒}了\\ `The student \underline{baffled} the math problem'\end{tabular} &
  ❌ \textbf{I}mpossible &
  ❌ \textbf{A}typical  &
  ✅ \textbf{R}elated & (IAR)\\
 &
  
  \begin{tabular}[c]{@{}l@{}}高材生把数学题\underline{困住}了\\ `The student \underline{restrained} the math problem'\end{tabular} &
  ❌ \textbf{I}mpossible &
  ❌ \textbf{A}typical  &
  ❌ \textbf{U}nrelated & (IAU)\\ \bottomrule
\end{tabular}\caption{Examples of the types of sentences used in the present study. All English sentences are drawn from \citet{vega-mendoza_2021_ConcurrentUseAnimacy}, and all Mandarin sentences are drawn from \citet{chow_2013_NoSemanticIllusions}.}
\label{tab:stims}
\end{table*}
\end{CJK*}

However, because this study was aimed at modeling human responses, it does not report the proportion of sentences where the possible continuation was assigned a higher probability than the impossible one (i.e., model accuracy). Thus, it is unclear how consistent this pattern is across individual items. Additionally, the original stimuli, as constructed by \citet{glenberg_2000_SymbolGroundingMeaning} explicitly balance the semantic relatedness of the afforded and unafforded options in order to ensure that it is not used as a heuristic---that is, to avoid cases where the afforded continuation has a higher semantic relatedness than the unafforded one. However, by the same token, this also removes the possibility of the reverse situation where reliance on a semantic relatedness heuristic may lead to mistakenly over-estimating the probability of semantically related words relative to unrelated words. Thus, the results may also over-estimate the extent to which language models can reliably predict possible events to be more likely than impossible ones.

\section{The Present Study}
As has been discussed, the effects of possibility, typicality, and contextual semantic relatedness on language model probability have each been investigated individually. However, these are all intertwined---all else being equal, we should expect that typical events are also possible, and that words describing typical and possible events are more likely to be related to their semantic contexts than words describing atypical and impossible events. In this study, therefore, we aim to disentangle the effect of each by investigating the effects on prediction of cases where these cues conflict.

To do this, our version of the Sherlock Holmes task uses a minimal pairs paradigm (see, e.g., \citealp{linzen_2016_AssessingAbilityLSTMs,marvin_2018_TargetedSyntacticEvaluation,warstadt_2019_NeuralNetworkAcceptability,warstadt_2020_BLiMPBenchmarkLinguistic}) where the language model is presented with two sentences that differ by a critical word that makes the sentence either possible or impossible. In order to allow any language model to carry out our task (not just large or finetuned models, see \citealp{hu_2024_AuxiliaryTaskDemandsa}), we follow previous work such as BLiMP \citep{warstadt_2020_BLiMPBenchmarkLinguistic} and directly compare the probability that a language model assigns to each sentence to determine the more probable one.

As shown in \autoref{tab:stims}, with each sentence, the critical word (underlined) is manipulated such that the event it describes is either Possible or Impossible and either Typical or Atypical. The critical word itself is either semantically Related or Unrelated to its preceding context. Thus, a minimal pair might pit a Possible-Typical-Related (PTR) sentence (e.g., \textit{the cure for the disease was discovered by the \textbf{doctor}}) against its corresponding Impossible-Atypical-Unrelated sentence (e.g., \textit{the cure for the disease was discovered by the \textbf{stamp}}). We consider a language model to be correct if it assigns a higher probability to the former (possible) sentence than the latter (impossible) sentence.

In this study, as in previous work comparing possible and impossible sentences in language models \citep{kauf_2023_EventKnowledgeLarge}, we look at impossibility arising from animacy violations. An animacy violation occurs when an inanimate entity has an event role that requires animacy. For example, in \textit{the cure for the disease was discovered by the...}, the discoverer needs to be animate, and since stamps are inanimate, completing the sentence with \textit{stamp} is an animacy violation. Thus, when a sentence with an animacy violation is interpreted literally, it refers to an impossible state of affairs.

All English stimuli were drawn from a human study carried out by \citeauthor{vega-mendoza_2021_ConcurrentUseAnimacy} (\citeyear{vega-mendoza_2021_ConcurrentUseAnimacy}; based on \citealp{paczynski_2012_MultipleInfluencesSemantic}), and all Mandarin stimuli from a study carried out by \citet{chow_2013_NoSemanticIllusions}. English minimal pairs are exactly as shown in \autoref{tab:stims}; while due to their nature, the Mandarin stimuli are embedded in larger sentences as they appear in \citet{chow_2013_NoSemanticIllusions}, which are identical across experimental conditions. All English tasks were made up of 154 sentence pairs, and all Mandarin tasks of 57 sentence pairs.

\begin{figure*}[t]
    \centering
    \includegraphics[width=\linewidth]{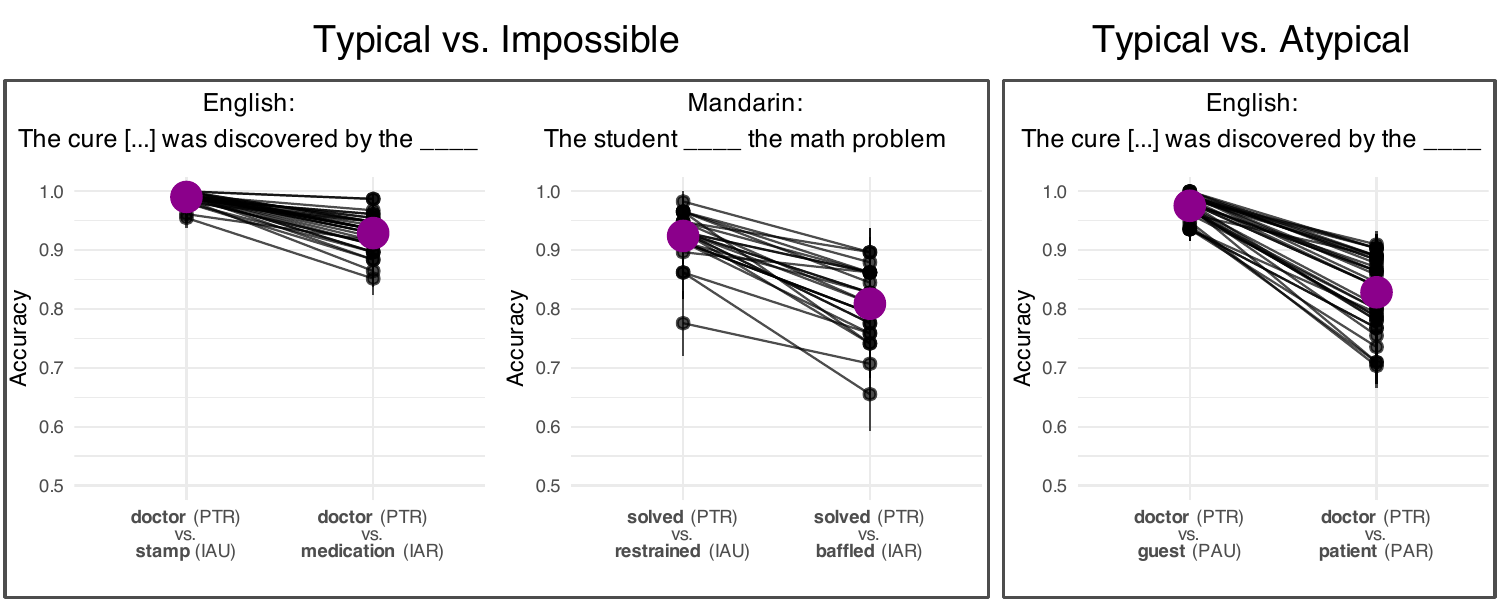}
    \caption{Language model scores for all tasks comparing sentences describing typical events (Possible-Typical-Related) to those describing impossible (Impossible-Atypical-Related or Impossible-Atypical-Unrelated) or merely atypical (Possible-Atypical-Related or Possible-Atypical-Unrelated) events. Individual model scores are shown in black (standard error calculated as in \citealp{biderman_2024_LessonsTrenchesReproducible}), while the mean over all models is shown in purple (standard error calculated over model means).}
    \label{fig:likely_events_eng}
\end{figure*}

\section{Experiment 1: Typical vs. Atypical}
\subsection{Introduction}
\label{ssec:plaus_vs_implaus_exp_intro}
In our first experiment, we first assess how well language models tell apart sentences denoting typical and atypical (both possible and impossible) events. Like \citet{kauf_2023_EventKnowledgeLarge}, we also test whether there is a difference in performance based on the type of atypical sentence compared (impossible vs. possible but atypical). We go beyond previous work both by looking at how performance is impacted by semantic relatedness, and by looking at a language other than English, namely, Mandarin.

\subsection{Method}
We construct minimal pairs based on the conditions described in \autoref{tab:stims}. For both languages, we test whether language models assign a higher probability to Possible-Typical-Related events compared to impossible critical words (see \autoref{tab:stims}). For English, we are also able to test whether language models assign a higher probability to Possible-Typical-Related sentences than to sentences with possible but atypical critical words.

We run our analyses using the Language Model Evaluation Harness \citep{gao_2021_FrameworkFewshotLanguage} on 35 language models of the BLOOM \citep{bigscienceworkshop_2023_BLOOM176BParameterOpenAccess}, Gemma \citep{gemmateam_2024_GemmaOpenModels,gemmateam_2024_Gemma2Improvinga}, Llama \citep{llamateam_2024_Llama3Herd}, mGPT \citep{shliazhko_2024_MGPTFewShotLearners}, Mistral \citep{jiang_2023_Mistral7B}, OLMo \citep{groeneveld_2024_OLMoAcceleratingScience}, Qwen \citep{yang_2024_Qwen2TechnicalReport,qwen_2025_Qwen25TechnicalReport}, SmolLM \citep{allal_2024_SmolLMBlazinglyFast}, XGLM \citep{lin_2022_FewshotLearningMultilingual}, and Yi \citep{01.ai_2025_YiOpenFoundation} model families (see \autoref{sec:appendix} for full list). Because we are interested in language models' world knowledge and the extent to which it is used in prediction---rather than their ability to answer questions about it---we limit our analysis to pretrained-only (or `base') models.

\begin{figure*}[t]
    \centering
    \includegraphics[width=\linewidth]{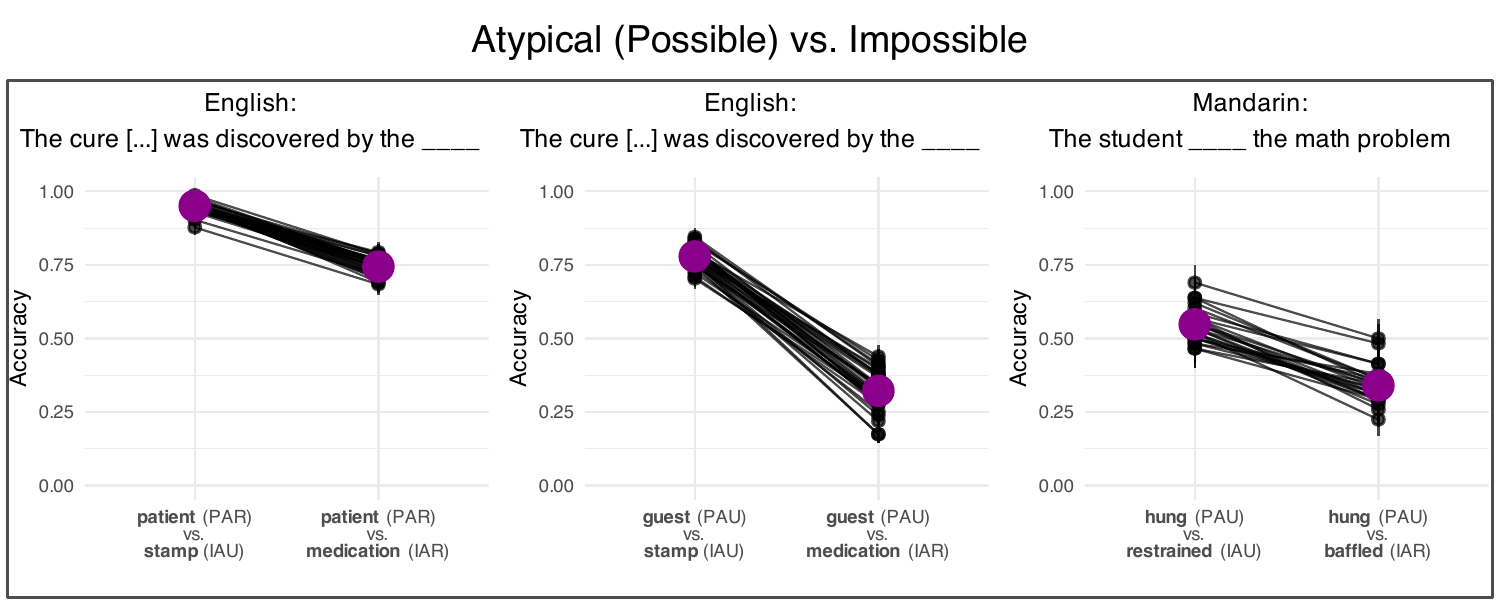}
    \caption{Language model scores for all tasks comparing sentences describing possible but atypical (Possible-Atypical-Related or Possible-Atypical-Unrelated) events to those describing impossible (Impossible-Atypical-Related or Impossible-Atypical-Unrelated) events. Individual model scores are shown in black, and overall mean scores in purple.}
    \label{fig:unlikely_events_eng}
\end{figure*}

\subsection{Results}

The results are shown in \autoref{fig:likely_events_eng} (the individual accuracy scores for all models is provided in \autoref{sec:appendix}). In addition to comparing the accuracies of different tasks numerically, we also run pairwise statistical tests. Specifically, to test whether there is a difference between language model performance at two different tasks, we use logistic mixed-effects regressions to predict whether a given answer is correct or not, and predict this based on which task it is, as well as including the maximal random effects structure that would converge with no singular fits, namely, random intercepts for each language model and sentence context. The specific statistical test was a likelihood ratio test including or excluding task as a predictor.

First, we see that on the typical vs. impossible tasks, accuracy is lower when the impossible critical word is related compared to when it is unrelated in both English ($\chi^2(1)=616.36,p<0.0001$) and Mandarin ($\chi^2(1)=163.06,p<0.0001$). The same is true with the typical vs. atypical comparison on English stimuli, where performance is lower when the implausible sentence is related rather than unrelated ($\chi^2(1)=1417.80,p<0.0001$). Finally, we replicate the result found by \citet{kauf_2023_EventKnowledgeLarge} that language models tend to be worse at the typical vs. atypical than the typical vs. impossible tasks. We find that this is the case both when the atypical or impossible critical word is related ($\chi^2(1)=587.32,p<0.0001$), and when it is unrelated ($\chi^2(1)=63.09,p<0.0001$). 

\subsection{Discussion}
In line with previous work \citep{kauf_2023_EventKnowledgeLarge}, our results show that language models can differentiate between possible and impossible events as well as typical and atypical events, and are better at the former comparison. 

In addition, we show for the first time that performance at each of these is affected by contextual semantic relatedness. If the atypical critical word (either possible or impossible) is semantically related to the event described in its context, it leads to decreased performance relative to when the atypical word is unrelated. This suggests that in cases where the atypical word is semantically related to the event in the context, there is an increased risk of it being calculated as more probable than the actually typical word than when the atypical word is semantically unrelated, including in cases where the atypical word renders the sentence impossible.

\section{Experiment 2: Atypical vs. Impossible}
\label{sec:exp_atypical_impossible}

\subsection{Introduction}
In Experiment 1, we showed that while language models can tell apart sentences denoting typical and atypical events, the extent to which this is the case is impacted by semantic relatedness, and specifically, the extent to which the critical word in a sentence is semantically related to its context. In Experiment 2, we instead focus on the more consequential question of how well language models can tell apart impossible and merely atypical events, again looking at how this is impacted by the semantic relatedness of the critical word to the context. However, in this experiment we evaluate models on the more difficult case where possible events are also atypical (Possible-Atypical-Unrelated or Possible-Atypical-Related).

\subsection{Method}
The method is the same as in Experiment 1, as are the impossible stimuli. However, for the possible stimuli, we instead draw on the atypical but possible sentence types: Possible-Atypical-Unrelated (English and Mandarin) and Possible-Atypical-Related (English only). We also limit our analysis to the subset of the language models used in Experiment 1 that were trained on Mandarin.

\subsection{Results}
The results are shown in \autoref{fig:unlikely_events_eng} (individual accuracy scores for all models provided in \autoref{sec:appendix}). First, we see that in all cases, an atypical but possible critical word leads to worse performance than a typical critical word.

There is a drastic drop in performance when the impossible critical word is related compared to when it is unrelated, both when the possible but atypical critical word is related (English only: $\chi^2(1)=1879.50,p<0.0001$) and when it is unrelated ($\chi^2(1)=3792.85,p<0.0001$; Mandarin\footnote{For this comparison, the null regression to be used in the likelihood ratio test did not converge, so we instead used the asymptotic Wald test to calculate the effect of task.}: $z =13.18,p<0.0001$). There is also a drop in performance on the English stimuli when the possible but atypical critical word is related compared to when it is unrelated (vs. Impossible-Atypical-Unrelated: $\chi^2(1)=1277.45,p<0.0001$; vs. Impossible-Atypical-Related: $\chi^2(1)=3570.05,p<0.0001$). When we compare performance to Experiment 1, we also see that for tasks where the possible critical word is related, the models perform significantly worse when the word is atypical rather than typical (vs. Impossible-Atypical-Unrelated: $\chi^2(1)=273.69,p<0.0001$; vs. Impossible-Atypical-Related: $\chi^2(1)=1240.70,p<0.0001$)

\subsection{Discussion}
These results show that language models are worse at differentiating between implausible and impossible events than they are between plausible and impossible events. While this may not be a surprising result given previous work \citep[e.g.,][]{jones_2022_DistrubutionalSemanticsStill,kauf_2023_EventKnowledgeLarge}, as far as we know, this is the first study to provide direct evidence of this.

In addition, this experiment highlights, perhaps even more strikingly than Experiment 1, the role of semantic relatedness in prediction. We see that performance is impacted not only by the extent to which the critical word of impossible sentences is related to the context, but also the extent to which the critical word of the possible sentences is related. In all cases, sentences with semantically related critical words are assigned higher probabilities.

Crucially, in the specific case where there is an event-unrelated possible critical word and event-related impossible word, we see that performance drops below chance in both languages---that is, the language models calculate the event-related impossible words to be more likely than the event-unrelated possible words more than half of the time.

\section{Experiment 3: Statistical Reliability}
\subsection{Introduction}
In addition to replicating \citet{kauf_2023_EventKnowledgeLarge}, we also make several novel findings in Experiments 1--2. Performance drops across the board when the possible critical word is atypical rather than typical, as well as when it is semantically unrelated rather than related. Additionally, performance drops when the impossible critical word is related rather than unrelated. While it may be unsurprising that language models appear to predict typical words to be more probable than atypical ones and semantically related words to be more probable than semantically unrelated words (especially given, e.g., \citealp{michaelov_2022_CollateralFacilitationHumans,michaelov_2024_StrongPredictionLanguage}), we show for the first time that this has a significant impact on their ability to assign higher probabilities to possible sentences than impossible ones.

\begin{figure*}[t]
    \centering
    \includegraphics[width=\linewidth]{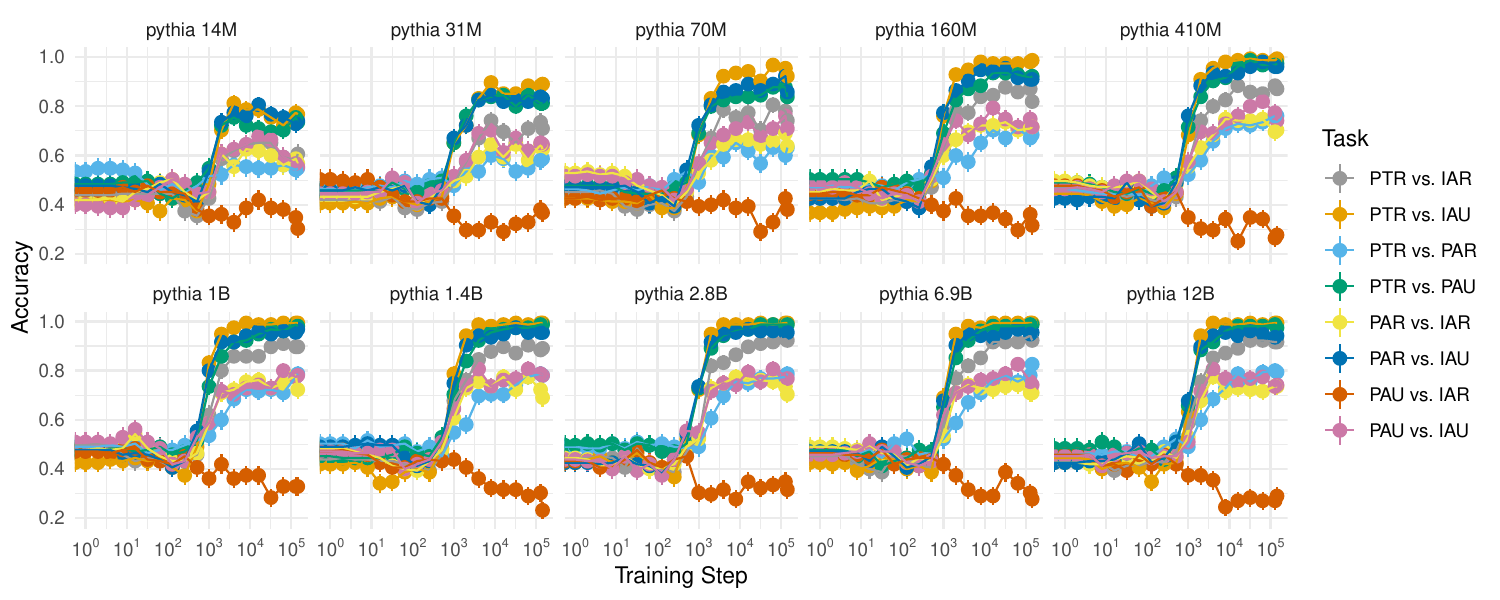}
    \caption{Pythia language model scores at all English tasks (from Experiments 1--2) over the course of training.}
    \label{fig:scaling}
\end{figure*}

\subsection{Method}
To investigate how robust this is and account for possible confounds, we carry out an analysis testing the extent to which the relatedness and typicality of each critical word impact language model performance. Because \citet{vega-mendoza_2021_ConcurrentUseAnimacy} provide both the numeric values of relatedness (calculated using Latent Semantic Analysis; \citealp{dumais_1988_UsingLatentSemantic}) and typicality (operationalized as human plausibility ratings), we use these, and thus only carry out our analyses on the English-language stimuli. We construct a logistic mixed-effects regression predicting whether a given language model correctly assigns the possible sentence a higher probability than the impossible one. As predictors, we include the semantic relatedness of the possible and impossible critical words, the typicality of the possible and impossible critical words, and the frequency of the possible and impossible critical words (a possible confound; see, e.g., \citealp{mccoy_2024_EmbersAutoregressionShow}). We also include random intercepts for each language model and sentence context, as well as random uncorrelated slopes of each predictor for each of these. Thus, we test the extent to which each predictor explains language model performance while accounting for the effects of the other predictors. We use likelihood ratio tests to compare the fit of regressions including or excluding the predictor of interest.

\subsection{Results and Discussion}
The results align with the results of Experiments 1--2. Specifically, we see that, for a given pair, the possible critical word being more semantically related improves performance ($\chi^2(1)=176.17,p<0.0001$), while the impossible critical word being more semantically related degrades performance ($\chi^2(1)=197.58,p<0.0001$). Similarly, we see that the possible critical word being more typical improves performance ( $\chi^2(1)=128.76,p<0.0001$) and the impossible word being more typical degrades performance ($\chi^2(1)=127.39,p<0.0001$). Thus, we see that plausibility and semantic relatedness do increase the probability assigned to a sentence by a language model to the extent that it significantly alters performance at the task. We also see that higher possible critical word frequency improves performance ($\chi^2(1)=	
394.32,p<0.0001$) while higher impossible critical word frequency degrades performance ($\chi^2(1)=302.54,p<0.0001$), suggesting that this is indeed a possible confound, and one that should be studied in future work.

\section{Experiment 4: Scaling Effects}
\label{sec:exp_scaling}
\subsection{Introduction}
Experiments 1 and 2 demonstrated fragility in language models' ability to differentiate between possible and impossible events, resulting from whether or not critical words in the sentences are related or unrelated to the context. However, there is a lot of variation between models. Given the general trend that larger models trained on more data tend to generally perform better \citep{kaplan_2020_ScalingLawsNeural,rae_2022_ScalingLanguageModels,wei_2022_EmergentAbilitiesLarge,hoffmann_2022_TrainingComputeOptimalLarge}, one might expect that the best-performing models on our tasks are the larger, more powerful models, and thus, that ultimately we may expect the relatedness effect on event understanding to disappear eventually with scale.

However, a preliminary examination of individual model performance at the Possible-Atypical-Unrelated vs. Impossible-Atypical-Related event comparison (see \autoref{tab:english_scores}) suggests that this is not the case. In fact, for some model families we see the opposite pattern: BLOOM 560M, 1B, and 1.7B are all better-performing than the 3B and 7B models, XGLM 564M out-performs other XGLM models, mGPT 1.3B out-performs mGPT 13B, and Gemma 1 models out-perform Gemma 2 models.

In this experiment, therefore, we take a systematic approach to investigating whether the performance of language models at the English tasks in Experiments 1 and 2 is correlated with their scale (i.e., number of parameters and number of training tokens). To do this, we turn to the Pythia models \citep{biderman_2023_PythiaSuiteAnalyzing}, a suite of language models of different sizes (in terms of number of parameters) that are provided at various checkpoints over the course of training. We investigate the patterns in model performance based on different numbers of parameters and over the course of training.

\subsection{Method}
The datasests are the same as those in Experiments 1 and 2. We use the 14M, 31M, 70M, 160M, 410M, 1B, 1.4B, 2.8B, 6.9B, and 12B parameter Pythia models \citep{biderman_2023_PythiaSuiteAnalyzing}, at training steps 0, 1, 2, 4, 8, 16, 32, 64, 128, 256, 512, 1000, 2000, 4000, 8000, 16000, 32000, 64000, 128000, and 143000 (the fully-trained model), where each training step comprises of $\sim$2M tokens.

\subsection{Results}
\autoref{fig:scaling} shows a striking difference between the Possible-Atypical-Unrelated vs. Impossible-Atypical-Related (\textit{the cure for the disease was discovered by the \textbf{guest}} vs. \textbf{\textit{medication}}) comparison and all other tasks. With the other tasks, there is a general pattern of improvement over the course of training, with the larger models showing numerically the best final scores. However, for the critical Possible-Atypical-Unrelated vs. Impossible-Atypical-Related comparison---where in Experiment 2, we saw at best chance accuracy across all models in both languages---performance never increases above chance.

\subsection{Discussion}
The effect of relatedness is not one that language models can fully `grow out of'. On the Possible-Atypical-Unrelated vs. Impossible-Atypical-Related task, there is little difference between larger and smaller models, and if anything, the performance of larger models deteriorates over the course of training. 

\section{General Discussion}
Previous work has shown that the predictions of language models are highly correlated with event typicality \citep{kauf_2023_EventKnowledgeLarge,michaelov_2024_StrongPredictionLanguage}. In our study, we similarly find that language models are sensitive to large differences in typicality, for example when distinguishing between impossible events and typical events. However, we also find that they perform significantly worse when distinguishing between impossible events and less-typical but still possible events. Thus, our results suggest that there is ample room for potential improvement at this end of the scale.

A larger and perhaps more striking result is the effect of semantic relatedness. A language model will become worse at distinguishing between a possible and impossible event when the possible event involves a critical word that is semantically unrelated to its preceding context, when the impossible critical word is semantically related, and especially when both of these are the case.

Combined with previous work showing that all else being equal, language models will often tend to predict that words that are semantically related to the context are more likely than unrelated ones \citep{misra_2020_ExploringBERTsSensitivity,michaelov_2022_CollateralFacilitationHumans}, these results suggest that language models at least in part rely on this relatedness (in addition to typicality) as a cue for which words are more probable. 

The presence of such a heuristic would be in line with other work showing that the strong performance of language models and other artificial intelligence systems can often be explained by them learning simpler `shortcuts' or other heuristics that correlate (often but not always) with the task at hand \citep[see, e.g.,][]{gururangan_2018_AnnotationArtifactsNatural,mccoy_2019_RightWrongReasons,mccoy_2024_EmbersAutoregressionShow,abdou_2020_SensitivityLanguageModels,geirhos_2020_ShortcutLearningDeep,schramowski_2020_MakingDeepNeural,shah_2020_PitfallsSimplicityBias,zhang_2020_WinoWhyDeepDiagnosis,du_2021_InterpretingMitigatingShortcut,du_2022_ShortcutLearningLarge,elazar_2021_MeasuringImprovingConsistency,kavumba_2021_LearningLearnBe,ye_2021_CaseStudyShortcut,stefanik_2022_MethodsEstimatingImproving}. 

As previously noted, such a surface-level heuristic is likely adaptive. Generally, words that are semantically related to their context are likely to be more appropriate sentence continuations than unrelated words---indeed, all the critical words in the typical sentences are semantically related to their context. This may at least in part explain why such a heuristic exists, not just in language models, as our study suggests, but also in humans \citep[see, e.g.,][]{chwilla_2007_ImmediateIntegrationNovel,parviz_2011_UsingLanguageModels,ettinger_2016_ModelingN400Amplitude,frank_2017_WordPredictabilitySemantic,broderick_2018_ElectrophysiologicalCorrelatesSemantic,uchida_2021_ModelOnlineTemporalSpatial}.

Crucially, this effect of contextual semantic relatedness does not disappear with increased model scale---in fact, while models trained on more data appear to become better at detecting larger differences in typicality, they also appear to increasingly rely on semantic relatedness to the context as a cue. Thus, this issue is not likely to disappear as language models continue to be trained on ever more data---in fact, it may get worse.

\section{Conclusions}
While contemporary language models can tell apart possible and impossible events \citep{jones_2022_DistrubutionalSemanticsStill,kauf_2023_EventKnowledgeLarge}, we find that this breaks down when the possible events are not highly typical, and when semantic relatedness is not a cue to event possibility. This suggests that much of language models' reported performance at reflecting event probability in previous work may be explained by language models' sensitivity to differences in typicality and contextual semantic relatedness.

\section*{Limitations}
\begin{CJK*}{UTF8}{gbsn}
Our study is limited in that it only looks at one type of impossibility---namely, animacy violations---but there are also many other ways in which a sentence could refer to an impossible event. Additionally, one possible issue with animacy-violating sentences, as \citet{vega-mendoza_2021_ConcurrentUseAnimacy} note, is that they can be read as the beginning of a plausible sentence or figuratively as a plausible sentence. The former issue may be somewhat addressed by the fact that we add a period or full stop character at the end of each sentence (i.e., `.' for English and `。' for Mandarin). Additionally, \citet{vega-mendoza_2021_ConcurrentUseAnimacy} explicitly address these issues by constructing their stimuli to avoid such interpretations, and further verify this by carrying out a norming study where experimental participants were asked to rate the plausibility of their sentences. They find that the impossible (i.e., animacy-violating sentences) were rated as significantly less plausible than possible sentences. Thus, we do not believe that this issue is likely to be of concern for the English stimuli. While \citet{chow_2013_NoSemanticIllusions} do not report such a norming process for the Mandarin stimuli, the fact that we see almost indetical patterns in both the English and Mandarin stimuli suggests that this is unlikely to be a confound in their stimuli either.
\end{CJK*}

A second limitation is that while we do expand our study beyond just English, the other language we include is Mandarin, which is another extremely widely-spoken high-resource language for which there are many high-quality language models. However, given that our study tests the limitations of such models, this is less of an issue---results of the kind that we find are more concerning for language models that otherwise show good performance and thus might be more likely to be erroneously trusted \citep[for discussion, see, e.g.,][]{bender_2021_DangersStochasticParrots,raji_2022_FallacyAIFunctionality}.

Finally, we note that our dataset sizes are small. Each English-language task only includes 154 pairs, and each Mandarin task only includes 57 pairs. In principle, this means that the results may not be as robust as with a larger dataset. One way in which we address this is by including error bars in our figures, following the recommendation of the Association for Computational Linguistics Rolling Review Responsible Natural Language Processing Research checklist (i.e., the ARR Responsible NLP Research checklist; \citealp{carpuat_2024_ARRResponsibleNLP}; based on \citealp{dodge_2019_ShowYourWorka,beygelzimer_2021_IntroducingNeurIPS2021,rogers_2021_JustWhatYou}), which give an indication of the level of uncertainty in the results (for further discussion, see also, e.g., \citealp{reimers_2017_ReportingScoreDistributions,henderson_2018_DeepReinforcementLearning,forde_2019_ScientificMethodSciencea,gundersen_2018_StateArtReproducibilitya,marie_2021_ScientificCredibilityMachinea,gundersen_2023_SourcesIrreproducibilityMachinea,kapoor_2024_REFORMSConsensusbasedRecommendationsa,biderman_2024_LessonsTrenchesReproducible}). Another is that the analyses reported in Experiment 3 on the English tasks are carried out at the item rather than task level, and thus also account for the number of experimental items.

\section*{Ethical Considerations}
We do not believe that our study raises any ethical concerns. In fact, we hope that in demonstrating the limitations of contemporary language models, we will increase the extent to which readers are careful in their use of such technologies.

From an environmental perspective, our study is of minimal impact. All experiments were carried out on a computing cluster on NVIDIA A100 GPUs in under 2 GPU hours.

Our use of all the language models for research purposes falls within their terms of use and license agreements, as does our use of the Language Model Evaluation Harness. The experimental stimuli from \citet{chow_2013_NoSemanticIllusions} and \citet{vega-mendoza_2021_ConcurrentUseAnimacy} are scientific research materials, and as such, we believe that their use for scientific research falls under the category of fair use. 

We provide all of our data, code and analyses at: \url{https://osf.io/r6xns/?view_only=0567164a44f64530bde24c3bc5f1ddbd}

\section*{Acknowledgments}
We would like to thank Wing-Yee Chow, Colin Phillips, Mariana Vega-Mendoza, Martin J. Pickering, Mante S. Nieuwland, Martin Paczynski, and Gina R. Kuperberg for making their experimental stimuli accessible. We would also like to thank the members of the Computational Psycholinguistics Laboratory at MIT and the Language and Cognition Laboratory at UCSD for their advice and discussion, in particular Cedegao E. Zhang, Roger Levy, and Catherine Arnett. James Michaelov was supported by a grant from the Andrew W. Mellon foundation (\#2210-13947) during the writing of this paper.

\bibliography{library}

\appendix

\section{Model Scores}
\label{sec:appendix}

We include the scores for all models. \autoref{tab:english_scores} shows the scores on all English-language tasks from Experiments 1 and 2 for all language models used in the analysis. \autoref{tab:mandarin_scores} shows the scores on all Mandarin tasks from Experiments 1 and 2 for all language models used in the analysis. The remaining tables (\autoref{tab:pythia_14m_score}--\autoref{tab:pythia_12b_score}) show the scores on all English-language tasks from Experiment 3 on all checkpoints (0, 1, 2, 4, 8, 16, 32, 64, 128, 256, 512, 1000, 2000, 4000, 8000, 16000, 32000, 64000, 128000, and 143000) of all the Pythia models (14M, 31M, 70M, 160M, 410M, 1B, 1.4B, 2.8B, 6.9B, and 12B) tested.

\begin{table*}[t]
\small
\centering
\begin{tabular}{@{}lcccccccc@{}}
\toprule
\textbf{Model} &
  \textbf{\begin{tabular}[c]{@{}c@{}}PTR\\ vs.\\ IAR\end{tabular}} &
  \textbf{\begin{tabular}[c]{@{}c@{}}PTR\\ vs.\\ IAU\end{tabular}} &
  \textbf{\begin{tabular}[c]{@{}c@{}}PTR\\ vs.\\ PAR\end{tabular}} &
  \textbf{\begin{tabular}[c]{@{}c@{}}PTR\\ vs.\\ PAU\end{tabular}} &
  \textbf{\begin{tabular}[c]{@{}c@{}}PAR\\ vs.\\ IAR\end{tabular}} &
  \textbf{\begin{tabular}[c]{@{}c@{}}PAR\\ vs.\\ IAU\end{tabular}} &
  \textbf{\begin{tabular}[c]{@{}c@{}}PAU\\ vs.\\ IAR\end{tabular}} &
  \textbf{\begin{tabular}[c]{@{}c@{}}PAU\\ vs.\\ IAU\end{tabular}} \\ \midrule
\texttt{01-ai/Yi-1.5-6B}            & 0.858 & 0.955 & 0.800 & 0.942 & 0.690 & 0.871 & 0.277 & 0.684 \\
\texttt{01-ai/Yi-1.5-9B}            & 0.961 & 0.994 & 0.877 & 0.987 & 0.710 & 0.948 & 0.226 & 0.761 \\
\texttt{01-ai/Yi-6B}                & 0.961 & 0.994 & 0.852 & 0.994 & 0.742 & 0.935 & 0.265 & 0.748 \\
\texttt{01-ai/Yi-9B}                & 0.948 & 0.994 & 0.871 & 0.987 & 0.729 & 0.961 & 0.310 & 0.723 \\
\texttt{HuggingFaceTB/SmolLM-1.7B}  & 0.884 & 0.987 & 0.832 & 0.981 & 0.710 & 0.987 & 0.277 & 0.703 \\
\texttt{HuggingFaceTB/SmolLM-135M}  & 0.903 & 0.987 & 0.748 & 0.948 & 0.787 & 0.974 & 0.355 & 0.774 \\
\texttt{HuggingFaceTB/SmolLM-360M}  & 0.923 & 0.974 & 0.819 & 0.987 & 0.768 & 0.974 & 0.252 & 0.729 \\
\texttt{Qwen/Qwen2-0.5B}            & 0.935 & 1.000 & 0.839 & 1.000 & 0.761 & 0.981 & 0.335 & 0.800 \\
\texttt{Qwen/Qwen2-1.5B}            & 0.942 & 1.000 & 0.832 & 0.994 & 0.761 & 0.961 & 0.348 & 0.813 \\
\texttt{Qwen/Qwen2-7B}              & 0.942 & 1.000 & 0.877 & 1.000 & 0.723 & 0.942 & 0.361 & 0.787 \\
\texttt{Qwen/Qwen2.5-0.5B}          & 0.935 & 0.994 & 0.800 & 0.981 & 0.781 & 0.987 & 0.381 & 0.839 \\
\texttt{Qwen/Qwen2.5-1.5B}          & 0.955 & 0.987 & 0.852 & 0.994 & 0.755 & 0.968 & 0.342 & 0.787 \\
\texttt{Qwen/Qwen2.5-14B}           & 0.974 & 1.000 & 0.903 & 0.994 & 0.794 & 0.948 & 0.387 & 0.774 \\
\texttt{Qwen/Qwen2.5-3B}            & 0.916 & 0.968 & 0.781 & 0.935 & 0.729 & 0.910 & 0.342 & 0.768 \\
\texttt{Qwen/Qwen2.5-7B}            & 0.955 & 1.000 & 0.890 & 0.994 & 0.761 & 0.955 & 0.400 & 0.781 \\
\texttt{ai-forever/mGPT}            & 0.865 & 0.987 & 0.710 & 0.935 & 0.755 & 0.942 & 0.368 & 0.774 \\
\texttt{ai-forever/mGPT-13B}        & 0.923 & 0.994 & 0.800 & 0.968 & 0.716 & 0.955 & 0.297 & 0.703 \\
\texttt{allenai/OLMo-2-1124-13B}    & 0.961 & 1.000 & 0.903 & 0.981 & 0.768 & 0.948 & 0.394 & 0.800 \\
\texttt{allenai/OLMo-2-1124-7B}     & 0.955 & 1.000 & 0.890 & 0.981 & 0.723 & 0.935 & 0.348 & 0.806 \\
\texttt{bigscience/bloom-1b1}       & 0.929 & 1.000 & 0.710 & 0.961 & 0.774 & 0.955 & 0.419 & 0.826 \\
\texttt{bigscience/bloom-1b7}       & 0.910 & 0.994 & 0.768 & 0.942 & 0.781 & 0.955 & 0.426 & 0.794 \\
\texttt{bigscience/bloom-3b}        & 0.903 & 0.987 & 0.800 & 0.974 & 0.755 & 0.961 & 0.355 & 0.748 \\
\texttt{bigscience/bloom-560m}      & 0.890 & 0.987 & 0.735 & 0.935 & 0.781 & 0.942 & 0.413 & 0.845 \\
\texttt{bigscience/bloom-7b1}       & 0.942 & 0.994 & 0.800 & 0.981 & 0.768 & 0.955 & 0.381 & 0.794 \\
\texttt{facebook/xglm-1.7B}         & 0.916 & 0.987 & 0.755 & 0.974 & 0.761 & 0.974 & 0.297 & 0.748 \\
\texttt{facebook/xglm-2.9B}         & 0.910 & 0.994 & 0.774 & 0.987 & 0.742 & 0.981 & 0.277 & 0.768 \\
\texttt{facebook/xglm-4.5B}         & 0.903 & 0.994 & 0.781 & 0.981 & 0.729 & 0.955 & 0.303 & 0.787 \\
\texttt{facebook/xglm-564M}         & 0.884 & 0.994 & 0.703 & 0.968 & 0.761 & 0.974 & 0.329 & 0.794 \\
\texttt{facebook/xglm-7.5B}         & 0.903 & 0.994 & 0.794 & 0.974 & 0.716 & 0.955 & 0.245 & 0.761 \\
\texttt{google/gemma-2-2b}          & 0.929 & 1.000 & 0.852 & 0.994 & 0.697 & 0.961 & 0.181 & 0.761 \\
\texttt{google/gemma-2-9b}          & 0.935 & 1.000 & 0.865 & 1.000 & 0.729 & 0.974 & 0.168 & 0.748 \\
\texttt{google/gemma-2b}            & 0.942 & 1.000 & 0.826 & 0.981 & 0.742 & 0.948 & 0.400 & 0.794 \\
\texttt{google/gemma-7b}            & 0.923 & 0.987 & 0.884 & 0.961 & 0.742 & 0.948 & 0.413 & 0.826 \\
\texttt{meta-llama/Llama-3.1-8B}    & 0.981 & 1.000 & 0.903 & 1.000 & 0.794 & 0.974 & 0.310 & 0.794 \\
\texttt{meta-llama/Llama-3.2-1B}    & 0.942 & 1.000 & 0.826 & 0.994 & 0.710 & 0.955 & 0.277 & 0.781 \\
\texttt{meta-llama/Llama-3.2-3B}    & 0.942 & 1.000 & 0.871 & 0.994 & 0.755 & 0.955 & 0.297 & 0.813 \\
\texttt{meta-llama/Meta-Llama-3-8B} & 0.981 & 1.000 & 0.884 & 1.000 & 0.761 & 0.955 & 0.310 & 0.794 \\
\texttt{mistralai/Mistral-7B-v0.1}  & 0.955 & 0.994 & 0.903 & 0.994 & 0.742 & 0.948 & 0.284 & 0.781 \\
\texttt{mistralai/Mistral-7B-v0.3}  & 0.948 & 1.000 & 0.890 & 0.994 & 0.742 & 0.942 & 0.290 & 0.774 \\
\texttt{mistralai/Mistral-Nemo-Base-2407} & 0.955   & 1.000 & 0.897 & 0.987 & 0.755 & 0.942 & 0.310  & 0.787       \\ \bottomrule
\end{tabular}
\caption{Scores on the English tasks for all models used in Experiments 1 and 2.}
\label{tab:english_scores}
\end{table*}

\begin{table*}[b]
\small
\centering
\begin{tabular}{@{}lcccc@{}}
\toprule
\textbf{Model} &
  \textbf{\begin{tabular}[c]{@{}c@{}}PTR\\ vs.\\ IAR\end{tabular}} &
  \textbf{\begin{tabular}[c]{@{}c@{}}PTR\\ vs.\\ IAU\end{tabular}} &
  \textbf{\begin{tabular}[c]{@{}c@{}}PAU\\ vs.\\ IAR\end{tabular}} &
  \textbf{\begin{tabular}[c]{@{}c@{}}PAU\\ vs.\\ IAU\end{tabular}} \\ \midrule
\texttt{01-ai/Yi-1.5-6B}                  & 0.759       & 0.914       & 0.500       & 0.690       \\
\texttt{01-ai/Yi-1.5-9B}                  & 0.897       & 0.983       & 0.362       & 0.621       \\
\texttt{01-ai/Yi-6B}                      & 0.845       & 0.966       & 0.414       & 0.586       \\
\texttt{01-ai/Yi-9B}                      & 0.879       & 0.966       & 0.362       & 0.603       \\
\texttt{Qwen/Qwen2-0.5B}                  & 0.776       & 0.931       & 0.414       & 0.569       \\
\texttt{Qwen/Qwen2-1.5B}                  & 0.862       & 0.931       & 0.310       & 0.500       \\
\texttt{Qwen/Qwen2-7B}                    & 0.897       & 0.948       & 0.345       & 0.552       \\
\texttt{Qwen/Qwen2.5-0.5B}                & 0.793       & 0.914       & 0.328       & 0.569       \\
\texttt{Qwen/Qwen2.5-1.5B}                & 0.862       & 0.897       & 0.345       & 0.638       \\
\texttt{Qwen/Qwen2.5-14B}                 & 0.828       & 0.931       & 0.362       & 0.500       \\
\texttt{Qwen/Qwen2.5-3B}                  & 0.707       & 0.776       & 0.483       & 0.638       \\
\texttt{Qwen/Qwen2.5-7B}                  & 0.862       & 0.931       & 0.293       & 0.517       \\
\texttt{ai-forever/mGPT}                  & 0.655       & 0.862       & 0.293       & 0.534       \\
\texttt{ai-forever/mGPT-13B}              & 0.741       & 0.966       & 0.328       & 0.500       \\
\texttt{bigscience/bloom-1b1}             & 0.828       & 0.931       & 0.293       & 0.552       \\
\texttt{bigscience/bloom-1b7}             & 0.810       & 0.931       & 0.224       & 0.517       \\
\texttt{bigscience/bloom-3b}              & 0.828       & 0.914       & 0.259       & 0.569       \\
\texttt{bigscience/bloom-560m}            & 0.793       & 0.914       & 0.345       & 0.534       \\
\texttt{bigscience/bloom-7b1}             & 0.862       & 0.966       & 0.310       & 0.517       \\
\texttt{facebook/xglm-1.7B}               & 0.810       & 0.948       & 0.276       & 0.534       \\
\texttt{facebook/xglm-2.9B}               & 0.776       & 0.931       & 0.310       & 0.552       \\
\texttt{facebook/xglm-4.5B}               & 0.793       & 0.914       & 0.345       & 0.466       \\
\texttt{facebook/xglm-564M}               & 0.759       & 0.862       & 0.328       & 0.483       \\
\texttt{facebook/xglm-7.5B}               & 0.741       & 0.931       & 0.293       & 0.466       \\
\texttt{mistralai/Mistral-Nemo-Base-2407} & 0.862       & 0.948       & 0.379       & 0.483 \\ \bottomrule
\end{tabular}
\caption{Scores on the Mandarin tasks for all models used in Experiments 1 and 2. Note that we only ran these analyses on language models that were reported to be trained on Mandarin.}
\label{tab:mandarin_scores}
\end{table*}

\begin{table*}[]
\small
\centering
\begin{tabular}{@{}lrcccccccc@{}}
\toprule
\textbf{Model} &
  \textbf{Step} &
  \textbf{\begin{tabular}[c]{@{}c@{}}PTR\\ vs.\\ IAR\end{tabular}} &
  \textbf{\begin{tabular}[c]{@{}c@{}}PTR\\ vs.\\ IAU\end{tabular}} &
  \textbf{\begin{tabular}[c]{@{}c@{}}PTR\\ vs.\\ PAR\end{tabular}} &
  \textbf{\begin{tabular}[c]{@{}c@{}}PTR\\ vs.\\ PAU\end{tabular}} &
  \textbf{\begin{tabular}[c]{@{}c@{}}PAR\\ vs.\\ IAR\end{tabular}} &
  \textbf{\begin{tabular}[c]{@{}c@{}}PAR\\ vs.\\ IAU\end{tabular}} &
  \textbf{\begin{tabular}[c]{@{}c@{}}PAU\\ vs.\\ IAR\end{tabular}} &
  \textbf{\begin{tabular}[c]{@{}c@{}}PAU\\ vs.\\ IAU\end{tabular}} \\ \midrule
\texttt{EleutherAI/pythia-14m}  & 0      & 0.452 & 0.419 & 0.535 & 0.484 & 0.419 & 0.477 & 0.445 & 0.406 \\
\texttt{EleutherAI/pythia-14m}  & 1      & 0.452 & 0.419 & 0.535 & 0.484 & 0.419 & 0.477 & 0.445 & 0.406 \\
\texttt{EleutherAI/pythia-14m}  & 2      & 0.452 & 0.419 & 0.548 & 0.484 & 0.419 & 0.477 & 0.445 & 0.400 \\
\texttt{EleutherAI/pythia-14m}  & 4      & 0.452 & 0.419 & 0.548 & 0.484 & 0.426 & 0.477 & 0.445 & 0.394 \\
\texttt{EleutherAI/pythia-14m}  & 8      & 0.452 & 0.413 & 0.542 & 0.490 & 0.413 & 0.471 & 0.458 & 0.394 \\
\texttt{EleutherAI/pythia-14m}  & 16     & 0.445 & 0.419 & 0.529 & 0.490 & 0.452 & 0.465 & 0.452 & 0.426 \\
\texttt{EleutherAI/pythia-14m}  & 32     & 0.445 & 0.406 & 0.484 & 0.497 & 0.465 & 0.452 & 0.477 & 0.439 \\
\texttt{EleutherAI/pythia-14m}  & 64     & 0.439 & 0.374 & 0.458 & 0.452 & 0.484 & 0.484 & 0.452 & 0.471 \\
\texttt{EleutherAI/pythia-14m}  & 128    & 0.439 & 0.432 & 0.484 & 0.426 & 0.503 & 0.452 & 0.452 & 0.503 \\
\texttt{EleutherAI/pythia-14m}  & 256    & 0.381 & 0.426 & 0.452 & 0.458 & 0.452 & 0.439 & 0.445 & 0.484 \\
\texttt{EleutherAI/pythia-14m}  & 512    & 0.355 & 0.374 & 0.477 & 0.490 & 0.413 & 0.419 & 0.381 & 0.445 \\
\texttt{EleutherAI/pythia-14m}  & 1000   & 0.426 & 0.490 & 0.510 & 0.548 & 0.452 & 0.535 & 0.355 & 0.439 \\
\texttt{EleutherAI/pythia-14m}  & 2000   & 0.581 & 0.703 & 0.523 & 0.723 & 0.606 & 0.742 & 0.361 & 0.606 \\
\texttt{EleutherAI/pythia-14m}  & 4000   & 0.613 & 0.806 & 0.548 & 0.755 & 0.555 & 0.781 & 0.329 & 0.613 \\
\texttt{EleutherAI/pythia-14m}  & 8000   & 0.632 & 0.787 & 0.542 & 0.723 & 0.619 & 0.755 & 0.387 & 0.645 \\
\texttt{EleutherAI/pythia-14m}  & 16000  & 0.658 & 0.787 & 0.542 & 0.716 & 0.632 & 0.806 & 0.413 & 0.677 \\
\texttt{EleutherAI/pythia-14m}  & 32000  & 0.652 & 0.755 & 0.542 & 0.697 & 0.606 & 0.768 & 0.387 & 0.671 \\
\texttt{EleutherAI/pythia-14m}  & 64000  & 0.600 & 0.723 & 0.555 & 0.703 & 0.574 & 0.755 & 0.374 & 0.594 \\
\texttt{EleutherAI/pythia-14m}  & 128000 & 0.581 & 0.774 & 0.535 & 0.755 & 0.587 & 0.735 & 0.342 & 0.568 \\
\texttt{EleutherAI/pythia-14m}  & 143000 & 0.606 & 0.755 & 0.548 & 0.735 & 0.561 & 0.742 & 0.297 & 0.561 \\ \bottomrule
\end{tabular}
\caption{Pythia 14M scores on the English-language tasks.}
\label{tab:pythia_14m_score}
\end{table*}

\begin{table*}
\small
\centering
\begin{tabular}{@{}lrcccccccc@{}}
\toprule
\textbf{Model} &
  \textbf{Step} &
  \textbf{\begin{tabular}[c]{@{}c@{}}PTR\\ vs.\\ IAR\end{tabular}} &
  \textbf{\begin{tabular}[c]{@{}c@{}}PTR\\ vs.\\ IAU\end{tabular}} &
  \textbf{\begin{tabular}[c]{@{}c@{}}PTR\\ vs.\\ PAR\end{tabular}} &
  \textbf{\begin{tabular}[c]{@{}c@{}}PTR\\ vs.\\ PAU\end{tabular}} &
  \textbf{\begin{tabular}[c]{@{}c@{}}PAR\\ vs.\\ IAR\end{tabular}} &
  \textbf{\begin{tabular}[c]{@{}c@{}}PAR\\ vs.\\ IAU\end{tabular}} &
  \textbf{\begin{tabular}[c]{@{}c@{}}PAU\\ vs.\\ IAR\end{tabular}} &
  \textbf{\begin{tabular}[c]{@{}c@{}}PAU\\ vs.\\ IAU\end{tabular}} \\ \midrule
\texttt{EleutherAI/pythia-31m}  & 0      & 0.419 & 0.406 & 0.465 & 0.465 & 0.426 & 0.452 & 0.503 & 0.445 \\
\texttt{EleutherAI/pythia-31m}  & 1      & 0.419 & 0.406 & 0.465 & 0.465 & 0.426 & 0.452 & 0.503 & 0.445 \\
\texttt{EleutherAI/pythia-31m}  & 2      & 0.413 & 0.406 & 0.458 & 0.458 & 0.426 & 0.452 & 0.497 & 0.445 \\
\texttt{EleutherAI/pythia-31m}  & 4      & 0.413 & 0.413 & 0.458 & 0.465 & 0.426 & 0.465 & 0.490 & 0.445 \\
\texttt{EleutherAI/pythia-31m}  & 8      & 0.413 & 0.406 & 0.471 & 0.465 & 0.426 & 0.452 & 0.503 & 0.452 \\
\texttt{EleutherAI/pythia-31m}  & 16     & 0.419 & 0.426 & 0.477 & 0.452 & 0.452 & 0.471 & 0.471 & 0.452 \\
\texttt{EleutherAI/pythia-31m}  & 32     & 0.432 & 0.432 & 0.471 & 0.484 & 0.465 & 0.458 & 0.439 & 0.510 \\
\texttt{EleutherAI/pythia-31m}  & 64     & 0.387 & 0.413 & 0.497 & 0.465 & 0.477 & 0.477 & 0.413 & 0.490 \\
\texttt{EleutherAI/pythia-31m}  & 128    & 0.387 & 0.400 & 0.484 & 0.452 & 0.406 & 0.413 & 0.413 & 0.406 \\
\texttt{EleutherAI/pythia-31m}  & 256    & 0.406 & 0.413 & 0.503 & 0.477 & 0.426 & 0.400 & 0.426 & 0.445 \\
\texttt{EleutherAI/pythia-31m}  & 512    & 0.413 & 0.432 & 0.503 & 0.497 & 0.452 & 0.471 & 0.426 & 0.452 \\
\texttt{EleutherAI/pythia-31m}  & 1000   & 0.484 & 0.652 & 0.484 & 0.652 & 0.490 & 0.671 & 0.355 & 0.516 \\
\texttt{EleutherAI/pythia-31m}  & 2000   & 0.574 & 0.723 & 0.535 & 0.761 & 0.516 & 0.723 & 0.303 & 0.587 \\
\texttt{EleutherAI/pythia-31m}  & 4000   & 0.671 & 0.832 & 0.529 & 0.832 & 0.600 & 0.832 & 0.290 & 0.690 \\
\texttt{EleutherAI/pythia-31m}  & 8000   & 0.742 & 0.897 & 0.600 & 0.832 & 0.645 & 0.845 & 0.329 & 0.697 \\
\texttt{EleutherAI/pythia-31m}  & 16000  & 0.697 & 0.852 & 0.587 & 0.839 & 0.600 & 0.813 & 0.297 & 0.600 \\
\texttt{EleutherAI/pythia-31m}  & 32000  & 0.645 & 0.852 & 0.535 & 0.800 & 0.652 & 0.832 & 0.323 & 0.671 \\
\texttt{EleutherAI/pythia-31m}  & 64000  & 0.703 & 0.890 & 0.548 & 0.845 & 0.581 & 0.819 & 0.329 & 0.619 \\
\texttt{EleutherAI/pythia-31m}  & 128000 & 0.742 & 0.884 & 0.574 & 0.806 & 0.632 & 0.839 & 0.374 & 0.639 \\
\texttt{EleutherAI/pythia-31m}  & 143000 & 0.716 & 0.890 & 0.581 & 0.806 & 0.619 & 0.832 & 0.381 & 0.639 \\ \bottomrule
\end{tabular}
\caption{Pythia 31M scores on the English-language tasks.}
\label{tab:pythia_31m_score}
\end{table*}

\begin{table*}
\small
\centering
\begin{tabular}{@{}lrcccccccc@{}}
\toprule
\textbf{Model} &
  \textbf{Step} &
  \textbf{\begin{tabular}[c]{@{}c@{}}PTR\\ vs.\\ IAR\end{tabular}} &
  \textbf{\begin{tabular}[c]{@{}c@{}}PTR\\ vs.\\ IAU\end{tabular}} &
  \textbf{\begin{tabular}[c]{@{}c@{}}PTR\\ vs.\\ PAR\end{tabular}} &
  \textbf{\begin{tabular}[c]{@{}c@{}}PTR\\ vs.\\ PAU\end{tabular}} &
  \textbf{\begin{tabular}[c]{@{}c@{}}PAR\\ vs.\\ IAR\end{tabular}} &
  \textbf{\begin{tabular}[c]{@{}c@{}}PAR\\ vs.\\ IAU\end{tabular}} &
  \textbf{\begin{tabular}[c]{@{}c@{}}PAU\\ vs.\\ IAR\end{tabular}} &
  \textbf{\begin{tabular}[c]{@{}c@{}}PAU\\ vs.\\ IAU\end{tabular}} \\ \midrule
\texttt{EleutherAI/pythia-70m}  & 0      & 0.445 & 0.426 & 0.477 & 0.477 & 0.529 & 0.471 & 0.432 & 0.516 \\
\texttt{EleutherAI/pythia-70m}  & 1      & 0.445 & 0.426 & 0.477 & 0.477 & 0.529 & 0.471 & 0.432 & 0.516 \\
\texttt{EleutherAI/pythia-70m}  & 2      & 0.445 & 0.426 & 0.465 & 0.477 & 0.535 & 0.471 & 0.439 & 0.529 \\
\texttt{EleutherAI/pythia-70m}  & 4      & 0.452 & 0.419 & 0.452 & 0.477 & 0.529 & 0.471 & 0.426 & 0.516 \\
\texttt{EleutherAI/pythia-70m}  & 8      & 0.445 & 0.413 & 0.465 & 0.477 & 0.523 & 0.471 & 0.432 & 0.516 \\
\texttt{EleutherAI/pythia-70m}  & 16     & 0.394 & 0.413 & 0.477 & 0.452 & 0.510 & 0.477 & 0.432 & 0.490 \\
\texttt{EleutherAI/pythia-70m}  & 32     & 0.381 & 0.413 & 0.465 & 0.445 & 0.477 & 0.452 & 0.419 & 0.503 \\
\texttt{EleutherAI/pythia-70m}  & 64     & 0.400 & 0.439 & 0.477 & 0.445 & 0.452 & 0.406 & 0.439 & 0.484 \\
\texttt{EleutherAI/pythia-70m}  & 128    & 0.406 & 0.394 & 0.458 & 0.445 & 0.439 & 0.439 & 0.458 & 0.471 \\
\texttt{EleutherAI/pythia-70m}  & 256    & 0.374 & 0.387 & 0.503 & 0.471 & 0.400 & 0.394 & 0.426 & 0.419 \\
\texttt{EleutherAI/pythia-70m}  & 512    & 0.432 & 0.477 & 0.529 & 0.497 & 0.452 & 0.542 & 0.413 & 0.458 \\
\texttt{EleutherAI/pythia-70m}  & 1000   & 0.568 & 0.684 & 0.497 & 0.690 & 0.535 & 0.723 & 0.394 & 0.574 \\
\texttt{EleutherAI/pythia-70m}  & 2000   & 0.671 & 0.832 & 0.600 & 0.800 & 0.581 & 0.806 & 0.413 & 0.645 \\
\texttt{EleutherAI/pythia-70m}  & 4000   & 0.781 & 0.923 & 0.594 & 0.826 & 0.639 & 0.858 & 0.419 & 0.684 \\
\texttt{EleutherAI/pythia-70m}  & 8000   & 0.755 & 0.935 & 0.652 & 0.832 & 0.671 & 0.865 & 0.394 & 0.710 \\
\texttt{EleutherAI/pythia-70m}  & 16000  & 0.774 & 0.942 & 0.619 & 0.839 & 0.677 & 0.890 & 0.394 & 0.735 \\
\texttt{EleutherAI/pythia-70m}  & 32000  & 0.703 & 0.903 & 0.581 & 0.845 & 0.639 & 0.865 & 0.290 & 0.677 \\
\texttt{EleutherAI/pythia-70m}  & 64000  & 0.806 & 0.968 & 0.632 & 0.884 & 0.671 & 0.897 & 0.329 & 0.710 \\
\texttt{EleutherAI/pythia-70m}  & 128000 & 0.781 & 0.955 & 0.594 & 0.871 & 0.690 & 0.923 & 0.439 & 0.761 \\
\texttt{EleutherAI/pythia-70m}  & 143000 & 0.729 & 0.923 & 0.632 & 0.832 & 0.645 & 0.858 & 0.381 & 0.710 \\ \bottomrule
\end{tabular}
\caption{Pythia 70M scores on the English-language tasks.}
\label{tab:pythia_70m_score}
\end{table*}

\begin{table*}
\small
\centering
\begin{tabular}{@{}lrcccccccc@{}}
\toprule
\textbf{Model} &
  \textbf{Step} &
  \textbf{\begin{tabular}[c]{@{}c@{}}PTR\\ vs.\\ IAR\end{tabular}} &
  \textbf{\begin{tabular}[c]{@{}c@{}}PTR\\ vs.\\ IAU\end{tabular}} &
  \textbf{\begin{tabular}[c]{@{}c@{}}PTR\\ vs.\\ PAR\end{tabular}} &
  \textbf{\begin{tabular}[c]{@{}c@{}}PTR\\ vs.\\ PAU\end{tabular}} &
  \textbf{\begin{tabular}[c]{@{}c@{}}PAR\\ vs.\\ IAR\end{tabular}} &
  \textbf{\begin{tabular}[c]{@{}c@{}}PAR\\ vs.\\ IAU\end{tabular}} &
  \textbf{\begin{tabular}[c]{@{}c@{}}PAU\\ vs.\\ IAR\end{tabular}} &
  \textbf{\begin{tabular}[c]{@{}c@{}}PAU\\ vs.\\ IAU\end{tabular}} \\ \midrule
\texttt{EleutherAI/pythia-160m} & 0      & 0.426 & 0.368 & 0.471 & 0.503 & 0.445 & 0.413 & 0.445 & 0.471 \\
\texttt{EleutherAI/pythia-160m} & 1      & 0.426 & 0.368 & 0.471 & 0.503 & 0.445 & 0.413 & 0.445 & 0.471 \\
\texttt{EleutherAI/pythia-160m} & 2      & 0.426 & 0.368 & 0.477 & 0.503 & 0.458 & 0.413 & 0.452 & 0.465 \\
\texttt{EleutherAI/pythia-160m} & 4      & 0.426 & 0.381 & 0.471 & 0.503 & 0.452 & 0.406 & 0.452 & 0.484 \\
\texttt{EleutherAI/pythia-160m} & 8      & 0.426 & 0.387 & 0.471 & 0.503 & 0.452 & 0.445 & 0.439 & 0.471 \\
\texttt{EleutherAI/pythia-160m} & 16     & 0.406 & 0.394 & 0.484 & 0.490 & 0.490 & 0.465 & 0.439 & 0.445 \\
\texttt{EleutherAI/pythia-160m} & 32     & 0.406 & 0.374 & 0.477 & 0.471 & 0.419 & 0.465 & 0.471 & 0.426 \\
\texttt{EleutherAI/pythia-160m} & 64     & 0.413 & 0.394 & 0.458 & 0.490 & 0.439 & 0.426 & 0.458 & 0.439 \\
\texttt{EleutherAI/pythia-160m} & 128    & 0.413 & 0.419 & 0.490 & 0.497 & 0.445 & 0.413 & 0.445 & 0.439 \\
\texttt{EleutherAI/pythia-160m} & 256    & 0.406 & 0.413 & 0.503 & 0.484 & 0.406 & 0.387 & 0.426 & 0.413 \\
\texttt{EleutherAI/pythia-160m} & 512    & 0.471 & 0.497 & 0.523 & 0.535 & 0.503 & 0.555 & 0.400 & 0.510 \\
\texttt{EleutherAI/pythia-160m} & 1000   & 0.619 & 0.761 & 0.516 & 0.690 & 0.606 & 0.774 & 0.374 & 0.645 \\
\texttt{EleutherAI/pythia-160m} & 2000   & 0.781 & 0.916 & 0.600 & 0.826 & 0.671 & 0.865 & 0.432 & 0.742 \\
\texttt{EleutherAI/pythia-160m} & 4000   & 0.774 & 0.955 & 0.568 & 0.871 & 0.703 & 0.903 & 0.355 & 0.690 \\
\texttt{EleutherAI/pythia-160m} & 8000   & 0.839 & 0.981 & 0.645 & 0.890 & 0.729 & 0.942 & 0.348 & 0.729 \\
\texttt{EleutherAI/pythia-160m} & 16000  & 0.845 & 0.974 & 0.697 & 0.935 & 0.723 & 0.942 & 0.368 & 0.794 \\
\texttt{EleutherAI/pythia-160m} & 32000  & 0.877 & 0.974 & 0.665 & 0.935 & 0.742 & 0.955 & 0.348 & 0.755 \\
\texttt{EleutherAI/pythia-160m} & 64000  & 0.865 & 0.974 & 0.697 & 0.935 & 0.723 & 0.916 & 0.297 & 0.710 \\
\texttt{EleutherAI/pythia-160m} & 128000 & 0.871 & 0.981 & 0.658 & 0.923 & 0.729 & 0.910 & 0.361 & 0.748 \\
\texttt{EleutherAI/pythia-160m} & 143000 & 0.819 & 0.987 & 0.684 & 0.923 & 0.723 & 0.910 & 0.329 & 0.716 \\ \bottomrule
\end{tabular}
\caption{Pythia 160M scores on the English-language tasks.}
\label{tab:pythia_160m_score}
\end{table*}

\begin{table*}
\small
\centering
\begin{tabular}{@{}lrcccccccc@{}}
\toprule
\textbf{Model} &
  \textbf{Step} &
  \textbf{\begin{tabular}[c]{@{}c@{}}PTR\\ vs.\\ IAR\end{tabular}} &
  \textbf{\begin{tabular}[c]{@{}c@{}}PTR\\ vs.\\ IAU\end{tabular}} &
  \textbf{\begin{tabular}[c]{@{}c@{}}PTR\\ vs.\\ PAR\end{tabular}} &
  \textbf{\begin{tabular}[c]{@{}c@{}}PTR\\ vs.\\ PAU\end{tabular}} &
  \textbf{\begin{tabular}[c]{@{}c@{}}PAR\\ vs.\\ IAR\end{tabular}} &
  \textbf{\begin{tabular}[c]{@{}c@{}}PAR\\ vs.\\ IAU\end{tabular}} &
  \textbf{\begin{tabular}[c]{@{}c@{}}PAU\\ vs.\\ IAR\end{tabular}} &
  \textbf{\begin{tabular}[c]{@{}c@{}}PAU\\ vs.\\ IAU\end{tabular}} \\ \midrule
\texttt{EleutherAI/pythia-410m} & 0      & 0.471 & 0.445 & 0.471 & 0.490 & 0.490 & 0.426 & 0.471 & 0.465 \\
\texttt{EleutherAI/pythia-410m} & 1      & 0.471 & 0.445 & 0.471 & 0.490 & 0.490 & 0.426 & 0.471 & 0.465 \\
\texttt{EleutherAI/pythia-410m} & 2      & 0.465 & 0.445 & 0.465 & 0.497 & 0.484 & 0.419 & 0.471 & 0.471 \\
\texttt{EleutherAI/pythia-410m} & 4      & 0.445 & 0.439 & 0.458 & 0.484 & 0.477 & 0.432 & 0.465 & 0.465 \\
\texttt{EleutherAI/pythia-410m} & 8      & 0.452 & 0.413 & 0.445 & 0.477 & 0.471 & 0.432 & 0.458 & 0.452 \\
\texttt{EleutherAI/pythia-410m} & 16     & 0.445 & 0.394 & 0.445 & 0.452 & 0.471 & 0.432 & 0.432 & 0.477 \\
\texttt{EleutherAI/pythia-410m} & 32     & 0.432 & 0.400 & 0.445 & 0.445 & 0.477 & 0.490 & 0.452 & 0.497 \\
\texttt{EleutherAI/pythia-410m} & 64     & 0.452 & 0.432 & 0.445 & 0.465 & 0.490 & 0.426 & 0.471 & 0.497 \\
\texttt{EleutherAI/pythia-410m} & 128    & 0.419 & 0.406 & 0.484 & 0.445 & 0.452 & 0.394 & 0.394 & 0.465 \\
\texttt{EleutherAI/pythia-410m} & 256    & 0.406 & 0.387 & 0.503 & 0.458 & 0.419 & 0.406 & 0.426 & 0.413 \\
\texttt{EleutherAI/pythia-410m} & 512    & 0.432 & 0.452 & 0.497 & 0.510 & 0.477 & 0.490 & 0.452 & 0.452 \\
\texttt{EleutherAI/pythia-410m} & 1000   & 0.561 & 0.684 & 0.523 & 0.716 & 0.555 & 0.761 & 0.348 & 0.581 \\
\texttt{EleutherAI/pythia-410m} & 2000   & 0.735 & 0.910 & 0.581 & 0.839 & 0.632 & 0.884 & 0.303 & 0.690 \\
\texttt{EleutherAI/pythia-410m} & 4000   & 0.800 & 0.955 & 0.658 & 0.897 & 0.684 & 0.935 & 0.297 & 0.690 \\
\texttt{EleutherAI/pythia-410m} & 8000   & 0.826 & 0.981 & 0.716 & 0.923 & 0.742 & 0.923 & 0.342 & 0.735 \\
\texttt{EleutherAI/pythia-410m} & 16000  & 0.890 & 0.994 & 0.729 & 0.968 & 0.742 & 0.935 & 0.252 & 0.761 \\
\texttt{EleutherAI/pythia-410m} & 32000  & 0.858 & 1.000 & 0.723 & 0.994 & 0.742 & 0.961 & 0.335 & 0.800 \\
\texttt{EleutherAI/pythia-410m} & 64000  & 0.865 & 0.994 & 0.742 & 0.974 & 0.748 & 0.981 & 0.342 & 0.819 \\
\texttt{EleutherAI/pythia-410m} & 128000 & 0.890 & 0.994 & 0.774 & 0.968 & 0.697 & 0.961 & 0.271 & 0.774 \\
\texttt{EleutherAI/pythia-410m} & 143000 & 0.877 & 1.000 & 0.774 & 0.981 & 0.703 & 0.961 & 0.284 & 0.742 \\ \bottomrule
\end{tabular}
\caption{Pythia 410M scores on the English-language tasks.}
\label{tab:pythia_410m_score}
\end{table*}

\begin{table*}
\small
\centering
\begin{tabular}{@{}lrcccccccc@{}}
\toprule
\textbf{Model} &
  \textbf{Step} &
  \textbf{\begin{tabular}[c]{@{}c@{}}PTR\\ vs.\\ IAR\end{tabular}} &
  \textbf{\begin{tabular}[c]{@{}c@{}}PTR\\ vs.\\ IAU\end{tabular}} &
  \textbf{\begin{tabular}[c]{@{}c@{}}PTR\\ vs.\\ PAR\end{tabular}} &
  \textbf{\begin{tabular}[c]{@{}c@{}}PTR\\ vs.\\ PAU\end{tabular}} &
  \textbf{\begin{tabular}[c]{@{}c@{}}PAR\\ vs.\\ IAR\end{tabular}} &
  \textbf{\begin{tabular}[c]{@{}c@{}}PAR\\ vs.\\ IAU\end{tabular}} &
  \textbf{\begin{tabular}[c]{@{}c@{}}PAU\\ vs.\\ IAR\end{tabular}} &
  \textbf{\begin{tabular}[c]{@{}c@{}}PAU\\ vs.\\ IAU\end{tabular}} \\ \midrule
\texttt{EleutherAI/pythia-1b}   & 0      & 0.503 & 0.426 & 0.490 & 0.477 & 0.484 & 0.477 & 0.471 & 0.516 \\
\texttt{EleutherAI/pythia-1b}   & 1      & 0.503 & 0.426 & 0.490 & 0.477 & 0.484 & 0.477 & 0.471 & 0.516 \\
\texttt{EleutherAI/pythia-1b}   & 2      & 0.510 & 0.426 & 0.490 & 0.477 & 0.490 & 0.477 & 0.471 & 0.516 \\
\texttt{EleutherAI/pythia-1b}   & 4      & 0.503 & 0.439 & 0.497 & 0.484 & 0.484 & 0.477 & 0.465 & 0.510 \\
\texttt{EleutherAI/pythia-1b}   & 8      & 0.465 & 0.432 & 0.490 & 0.477 & 0.484 & 0.458 & 0.477 & 0.523 \\
\texttt{EleutherAI/pythia-1b}   & 16     & 0.439 & 0.452 & 0.510 & 0.477 & 0.535 & 0.465 & 0.484 & 0.561 \\
\texttt{EleutherAI/pythia-1b}   & 32     & 0.445 & 0.432 & 0.471 & 0.477 & 0.484 & 0.452 & 0.439 & 0.510 \\
\texttt{EleutherAI/pythia-1b}   & 64     & 0.432 & 0.439 & 0.477 & 0.490 & 0.452 & 0.439 & 0.432 & 0.497 \\
\texttt{EleutherAI/pythia-1b}   & 128    & 0.406 & 0.406 & 0.477 & 0.452 & 0.432 & 0.406 & 0.419 & 0.419 \\
\texttt{EleutherAI/pythia-1b}   & 256    & 0.439 & 0.374 & 0.497 & 0.471 & 0.432 & 0.432 & 0.452 & 0.432 \\
\texttt{EleutherAI/pythia-1b}   & 512    & 0.458 & 0.490 & 0.516 & 0.523 & 0.458 & 0.561 & 0.406 & 0.529 \\
\texttt{EleutherAI/pythia-1b}   & 1000   & 0.619 & 0.826 & 0.535 & 0.735 & 0.587 & 0.800 & 0.368 & 0.581 \\
\texttt{EleutherAI/pythia-1b}   & 2000   & 0.794 & 0.948 & 0.600 & 0.858 & 0.710 & 0.916 & 0.406 & 0.723 \\
\texttt{EleutherAI/pythia-1b}   & 4000   & 0.865 & 0.974 & 0.684 & 0.916 & 0.716 & 0.916 & 0.361 & 0.710 \\
\texttt{EleutherAI/pythia-1b}   & 8000   & 0.858 & 1.000 & 0.723 & 0.929 & 0.755 & 0.942 & 0.374 & 0.729 \\
\texttt{EleutherAI/pythia-1b}   & 16000  & 0.871 & 0.994 & 0.716 & 0.955 & 0.755 & 0.961 & 0.361 & 0.748 \\
\texttt{EleutherAI/pythia-1b}   & 32000  & 0.897 & 0.987 & 0.723 & 0.968 & 0.729 & 0.948 & 0.284 & 0.742 \\
\texttt{EleutherAI/pythia-1b}   & 64000  & 0.916 & 1.000 & 0.710 & 0.968 & 0.748 & 0.968 & 0.323 & 0.800 \\
\texttt{EleutherAI/pythia-1b}   & 128000 & 0.903 & 1.000 & 0.774 & 0.981 & 0.781 & 0.961 & 0.335 & 0.774 \\
\texttt{EleutherAI/pythia-1b}   & 143000 & 0.903 & 1.000 & 0.794 & 0.981 & 0.723 & 0.974 & 0.335 & 0.781 \\ \bottomrule
\end{tabular}
\caption{Pythia 1B scores on the English-language tasks.}
\label{tab:pythia_1b_score}
\end{table*}

\begin{table*}
\small
\centering
\begin{tabular}{@{}lrcccccccc@{}}
\toprule
\textbf{Model} &
  \textbf{Step} &
  \textbf{\begin{tabular}[c]{@{}c@{}}PTR\\ vs.\\ IAR\end{tabular}} &
  \textbf{\begin{tabular}[c]{@{}c@{}}PTR\\ vs.\\ IAU\end{tabular}} &
  \textbf{\begin{tabular}[c]{@{}c@{}}PTR\\ vs.\\ PAR\end{tabular}} &
  \textbf{\begin{tabular}[c]{@{}c@{}}PTR\\ vs.\\ PAU\end{tabular}} &
  \textbf{\begin{tabular}[c]{@{}c@{}}PAR\\ vs.\\ IAR\end{tabular}} &
  \textbf{\begin{tabular}[c]{@{}c@{}}PAR\\ vs.\\ IAU\end{tabular}} &
  \textbf{\begin{tabular}[c]{@{}c@{}}PAU\\ vs.\\ IAR\end{tabular}} &
  \textbf{\begin{tabular}[c]{@{}c@{}}PAU\\ vs.\\ IAU\end{tabular}} \\ \midrule
\texttt{EleutherAI/pythia-1.4b} & 0      & 0.432 & 0.419 & 0.503 & 0.452 & 0.484 & 0.497 & 0.458 & 0.465 \\
\texttt{EleutherAI/pythia-1.4b} & 1      & 0.432 & 0.419 & 0.503 & 0.452 & 0.484 & 0.497 & 0.458 & 0.465 \\
\texttt{EleutherAI/pythia-1.4b} & 2      & 0.439 & 0.419 & 0.503 & 0.445 & 0.484 & 0.497 & 0.471 & 0.465 \\
\texttt{EleutherAI/pythia-1.4b} & 4      & 0.432 & 0.426 & 0.484 & 0.445 & 0.471 & 0.503 & 0.458 & 0.452 \\
\texttt{EleutherAI/pythia-1.4b} & 8      & 0.439 & 0.406 & 0.510 & 0.445 & 0.452 & 0.510 & 0.452 & 0.458 \\
\texttt{EleutherAI/pythia-1.4b} & 16     & 0.432 & 0.342 & 0.484 & 0.439 & 0.452 & 0.497 & 0.426 & 0.477 \\
\texttt{EleutherAI/pythia-1.4b} & 32     & 0.432 & 0.355 & 0.497 & 0.458 & 0.413 & 0.497 & 0.413 & 0.471 \\
\texttt{EleutherAI/pythia-1.4b} & 64     & 0.413 & 0.394 & 0.503 & 0.490 & 0.400 & 0.445 & 0.406 & 0.400 \\
\texttt{EleutherAI/pythia-1.4b} & 128    & 0.400 & 0.400 & 0.471 & 0.465 & 0.394 & 0.413 & 0.426 & 0.419 \\
\texttt{EleutherAI/pythia-1.4b} & 256    & 0.387 & 0.387 & 0.490 & 0.452 & 0.432 & 0.432 & 0.445 & 0.439 \\
\texttt{EleutherAI/pythia-1.4b} & 512    & 0.452 & 0.484 & 0.452 & 0.529 & 0.477 & 0.516 & 0.426 & 0.503 \\
\texttt{EleutherAI/pythia-1.4b} & 1000   & 0.665 & 0.781 & 0.548 & 0.716 & 0.606 & 0.742 & 0.426 & 0.652 \\
\texttt{EleutherAI/pythia-1.4b} & 2000   & 0.761 & 0.942 & 0.581 & 0.839 & 0.735 & 0.903 & 0.406 & 0.742 \\
\texttt{EleutherAI/pythia-1.4b} & 4000   & 0.845 & 0.981 & 0.690 & 0.923 & 0.761 & 0.942 & 0.368 & 0.806 \\
\texttt{EleutherAI/pythia-1.4b} & 8000   & 0.877 & 0.987 & 0.710 & 0.961 & 0.735 & 0.935 & 0.316 & 0.729 \\
\texttt{EleutherAI/pythia-1.4b} & 16000  & 0.903 & 0.994 & 0.710 & 0.968 & 0.768 & 0.955 & 0.323 & 0.768 \\
\texttt{EleutherAI/pythia-1.4b} & 32000  & 0.884 & 1.000 & 0.742 & 0.981 & 0.723 & 0.974 & 0.323 & 0.768 \\
\texttt{EleutherAI/pythia-1.4b} & 64000  & 0.910 & 0.994 & 0.781 & 0.981 & 0.768 & 0.961 & 0.303 & 0.800 \\
\texttt{EleutherAI/pythia-1.4b} & 128000 & 0.890 & 1.000 & 0.787 & 0.994 & 0.723 & 0.961 & 0.323 & 0.781 \\
\texttt{EleutherAI/pythia-1.4b} & 143000 & 0.897 & 1.000 & 0.781 & 0.994 & 0.690 & 0.961 & 0.252 & 0.774 \\ \bottomrule
\end{tabular}
\caption{Pythia 1.4B scores on the English-language tasks.}
\label{tab:pythia_1b4_score}
\end{table*}

\begin{table*}
\small
\centering
\begin{tabular}{@{}lrcccccccc@{}}
\toprule
\textbf{Model} &
  \textbf{Step} &
  \textbf{\begin{tabular}[c]{@{}c@{}}PTR\\ vs.\\ IAR\end{tabular}} &
  \textbf{\begin{tabular}[c]{@{}c@{}}PTR\\ vs.\\ IAU\end{tabular}} &
  \textbf{\begin{tabular}[c]{@{}c@{}}PTR\\ vs.\\ PAR\end{tabular}} &
  \textbf{\begin{tabular}[c]{@{}c@{}}PTR\\ vs.\\ PAU\end{tabular}} &
  \textbf{\begin{tabular}[c]{@{}c@{}}PAR\\ vs.\\ IAR\end{tabular}} &
  \textbf{\begin{tabular}[c]{@{}c@{}}PAR\\ vs.\\ IAU\end{tabular}} &
  \textbf{\begin{tabular}[c]{@{}c@{}}PAU\\ vs.\\ IAR\end{tabular}} &
  \textbf{\begin{tabular}[c]{@{}c@{}}PAU\\ vs.\\ IAU\end{tabular}} \\ \midrule
\texttt{EleutherAI/pythia-2.8b} & 0      & 0.439 & 0.419 & 0.490 & 0.497 & 0.439 & 0.426 & 0.426 & 0.445 \\
\texttt{EleutherAI/pythia-2.8b} & 1      & 0.439 & 0.419 & 0.490 & 0.497 & 0.439 & 0.426 & 0.426 & 0.445 \\
\texttt{EleutherAI/pythia-2.8b} & 2      & 0.439 & 0.432 & 0.484 & 0.503 & 0.439 & 0.432 & 0.426 & 0.439 \\
\texttt{EleutherAI/pythia-2.8b} & 4      & 0.452 & 0.452 & 0.484 & 0.497 & 0.419 & 0.419 & 0.406 & 0.439 \\
\texttt{EleutherAI/pythia-2.8b} & 8      & 0.432 & 0.426 & 0.490 & 0.503 & 0.406 & 0.426 & 0.406 & 0.400 \\
\texttt{EleutherAI/pythia-2.8b} & 16     & 0.394 & 0.439 & 0.477 & 0.477 & 0.452 & 0.445 & 0.439 & 0.452 \\
\texttt{EleutherAI/pythia-2.8b} & 32     & 0.484 & 0.458 & 0.510 & 0.503 & 0.419 & 0.413 & 0.477 & 0.394 \\
\texttt{EleutherAI/pythia-2.8b} & 64     & 0.419 & 0.426 & 0.497 & 0.497 & 0.400 & 0.413 & 0.426 & 0.413 \\
\texttt{EleutherAI/pythia-2.8b} & 128    & 0.400 & 0.406 & 0.503 & 0.471 & 0.406 & 0.381 & 0.406 & 0.374 \\
\texttt{EleutherAI/pythia-2.8b} & 256    & 0.387 & 0.368 & 0.490 & 0.471 & 0.413 & 0.439 & 0.439 & 0.439 \\
\texttt{EleutherAI/pythia-2.8b} & 512    & 0.477 & 0.523 & 0.484 & 0.516 & 0.484 & 0.542 & 0.445 & 0.555 \\
\texttt{EleutherAI/pythia-2.8b} & 1000   & 0.529 & 0.716 & 0.490 & 0.735 & 0.529 & 0.729 & 0.303 & 0.548 \\
\texttt{EleutherAI/pythia-2.8b} & 2000   & 0.819 & 0.948 & 0.600 & 0.884 & 0.729 & 0.916 & 0.297 & 0.716 \\
\texttt{EleutherAI/pythia-2.8b} & 4000   & 0.832 & 0.987 & 0.690 & 0.935 & 0.748 & 0.955 & 0.316 & 0.761 \\
\texttt{EleutherAI/pythia-2.8b} & 8000   & 0.871 & 0.994 & 0.761 & 0.968 & 0.781 & 0.955 & 0.277 & 0.742 \\
\texttt{EleutherAI/pythia-2.8b} & 16000  & 0.903 & 0.994 & 0.748 & 0.974 & 0.768 & 0.968 & 0.355 & 0.800 \\
\texttt{EleutherAI/pythia-2.8b} & 32000  & 0.910 & 1.000 & 0.774 & 0.994 & 0.748 & 0.961 & 0.323 & 0.761 \\
\texttt{EleutherAI/pythia-2.8b} & 64000  & 0.923 & 0.994 & 0.768 & 0.994 & 0.755 & 0.968 & 0.348 & 0.794 \\
\texttt{EleutherAI/pythia-2.8b} & 128000 & 0.935 & 0.994 & 0.781 & 0.994 & 0.735 & 0.955 & 0.355 & 0.774 \\
\texttt{EleutherAI/pythia-2.8b} & 143000 & 0.929 & 0.994 & 0.794 & 0.994 & 0.703 & 0.955 & 0.323 & 0.768 \\ \bottomrule
\end{tabular}
\caption{Pythia 2.8B scores on the English-language tasks.}
\label{tab:pythia_2b8_score}
\end{table*}

\begin{table*}
\small
\centering
\begin{tabular}{@{}lrcccccccc@{}}
\toprule
\textbf{Model} &
  \textbf{Step} &
  \textbf{\begin{tabular}[c]{@{}c@{}}PTR\\ vs.\\ IAR\end{tabular}} &
  \textbf{\begin{tabular}[c]{@{}c@{}}PTR\\ vs.\\ IAU\end{tabular}} &
  \textbf{\begin{tabular}[c]{@{}c@{}}PTR\\ vs.\\ PAR\end{tabular}} &
  \textbf{\begin{tabular}[c]{@{}c@{}}PTR\\ vs.\\ PAU\end{tabular}} &
  \textbf{\begin{tabular}[c]{@{}c@{}}PAR\\ vs.\\ IAR\end{tabular}} &
  \textbf{\begin{tabular}[c]{@{}c@{}}PAR\\ vs.\\ IAU\end{tabular}} &
  \textbf{\begin{tabular}[c]{@{}c@{}}PAU\\ vs.\\ IAR\end{tabular}} &
  \textbf{\begin{tabular}[c]{@{}c@{}}PAU\\ vs.\\ IAU\end{tabular}} \\ \midrule
\texttt{EleutherAI/pythia-6.9b} & 0      & 0.445 & 0.419 & 0.477 & 0.477 & 0.490 & 0.477 & 0.465 & 0.465 \\
\texttt{EleutherAI/pythia-6.9b} & 1      & 0.445 & 0.419 & 0.477 & 0.477 & 0.490 & 0.477 & 0.465 & 0.465 \\
\texttt{EleutherAI/pythia-6.9b} & 2      & 0.445 & 0.413 & 0.477 & 0.477 & 0.490 & 0.477 & 0.465 & 0.465 \\
\texttt{EleutherAI/pythia-6.9b} & 4      & 0.445 & 0.419 & 0.484 & 0.477 & 0.484 & 0.477 & 0.458 & 0.471 \\
\texttt{EleutherAI/pythia-6.9b} & 8      & 0.413 & 0.394 & 0.465 & 0.452 & 0.477 & 0.471 & 0.458 & 0.477 \\
\texttt{EleutherAI/pythia-6.9b} & 16     & 0.400 & 0.406 & 0.484 & 0.465 & 0.497 & 0.432 & 0.445 & 0.497 \\
\texttt{EleutherAI/pythia-6.9b} & 32     & 0.387 & 0.445 & 0.445 & 0.413 & 0.471 & 0.484 & 0.477 & 0.477 \\
\texttt{EleutherAI/pythia-6.9b} & 64     & 0.458 & 0.439 & 0.503 & 0.432 & 0.458 & 0.452 & 0.426 & 0.471 \\
\texttt{EleutherAI/pythia-6.9b} & 128    & 0.445 & 0.400 & 0.523 & 0.445 & 0.419 & 0.406 & 0.439 & 0.426 \\
\texttt{EleutherAI/pythia-6.9b} & 256    & 0.419 & 0.413 & 0.465 & 0.477 & 0.445 & 0.419 & 0.445 & 0.413 \\
\texttt{EleutherAI/pythia-6.9b} & 512    & 0.400 & 0.387 & 0.465 & 0.471 & 0.432 & 0.452 & 0.445 & 0.432 \\
\texttt{EleutherAI/pythia-6.9b} & 1000   & 0.574 & 0.690 & 0.510 & 0.652 & 0.542 & 0.690 & 0.419 & 0.619 \\
\texttt{EleutherAI/pythia-6.9b} & 2000   & 0.761 & 0.948 & 0.581 & 0.852 & 0.703 & 0.903 & 0.374 & 0.723 \\
\texttt{EleutherAI/pythia-6.9b} & 4000   & 0.826 & 0.981 & 0.671 & 0.923 & 0.742 & 0.948 & 0.316 & 0.742 \\
\texttt{EleutherAI/pythia-6.9b} & 8000   & 0.858 & 0.981 & 0.710 & 0.968 & 0.735 & 0.942 & 0.290 & 0.755 \\
\texttt{EleutherAI/pythia-6.9b} & 16000  & 0.916 & 1.000 & 0.768 & 0.981 & 0.723 & 0.942 & 0.290 & 0.768 \\
\texttt{EleutherAI/pythia-6.9b} & 32000  & 0.923 & 1.000 & 0.787 & 0.987 & 0.735 & 0.961 & 0.381 & 0.781 \\
\texttt{EleutherAI/pythia-6.9b} & 64000  & 0.923 & 1.000 & 0.748 & 0.987 & 0.742 & 0.968 & 0.335 & 0.819 \\
\texttt{EleutherAI/pythia-6.9b} & 128000 & 0.942 & 0.994 & 0.768 & 0.994 & 0.716 & 0.968 & 0.310 & 0.755 \\
\texttt{EleutherAI/pythia-6.9b} & 143000 & 0.929 & 0.994 & 0.813 & 0.994 & 0.710 & 0.961 & 0.284 & 0.748 \\ \bottomrule
\end{tabular}
\caption{Pythia 6.9B scores on the English-language tasks.}
\label{tab:pythia_6b9_score}
\end{table*}

\begin{table*}
\small
\centering
\begin{tabular}{@{}lrcccccccc@{}}
\toprule
\textbf{Model} &
  \textbf{Step} &
  \textbf{\begin{tabular}[c]{@{}c@{}}PTR\\ vs.\\ IAR\end{tabular}} &
  \textbf{\begin{tabular}[c]{@{}c@{}}PTR\\ vs.\\ IAU\end{tabular}} &
  \textbf{\begin{tabular}[c]{@{}c@{}}PTR\\ vs.\\ PAR\end{tabular}} &
  \textbf{\begin{tabular}[c]{@{}c@{}}PTR\\ vs.\\ PAU\end{tabular}} &
  \textbf{\begin{tabular}[c]{@{}c@{}}PAR\\ vs.\\ IAR\end{tabular}} &
  \textbf{\begin{tabular}[c]{@{}c@{}}PAR\\ vs.\\ IAU\end{tabular}} &
  \textbf{\begin{tabular}[c]{@{}c@{}}PAU\\ vs.\\ IAR\end{tabular}} &
  \textbf{\begin{tabular}[c]{@{}c@{}}PAU\\ vs.\\ IAU\end{tabular}} \\ \midrule
\texttt{EleutherAI/pythia-12b}  & 0      & 0.426 & 0.439 & 0.471 & 0.490 & 0.419 & 0.426 & 0.458 & 0.458 \\
\texttt{EleutherAI/pythia-12b}  & 1      & 0.426 & 0.439 & 0.471 & 0.490 & 0.419 & 0.426 & 0.458 & 0.458 \\
\texttt{EleutherAI/pythia-12b}  & 2      & 0.419 & 0.439 & 0.471 & 0.490 & 0.426 & 0.426 & 0.458 & 0.458 \\
\texttt{EleutherAI/pythia-12b}  & 4      & 0.426 & 0.432 & 0.477 & 0.484 & 0.419 & 0.432 & 0.458 & 0.458 \\
\texttt{EleutherAI/pythia-12b}  & 8      & 0.419 & 0.400 & 0.452 & 0.510 & 0.419 & 0.432 & 0.413 & 0.419 \\
\texttt{EleutherAI/pythia-12b}  & 16     & 0.400 & 0.419 & 0.471 & 0.490 & 0.445 & 0.439 & 0.419 & 0.452 \\
\texttt{EleutherAI/pythia-12b}  & 32     & 0.413 & 0.426 & 0.471 & 0.458 & 0.452 & 0.465 & 0.394 & 0.465 \\
\texttt{EleutherAI/pythia-12b}  & 64     & 0.413 & 0.406 & 0.445 & 0.484 & 0.439 & 0.503 & 0.458 & 0.471 \\
\texttt{EleutherAI/pythia-12b}  & 128    & 0.432 & 0.348 & 0.452 & 0.477 & 0.426 & 0.426 & 0.419 & 0.445 \\
\texttt{EleutherAI/pythia-12b}  & 256    & 0.426 & 0.419 & 0.490 & 0.439 & 0.471 & 0.452 & 0.458 & 0.452 \\
\texttt{EleutherAI/pythia-12b}  & 512    & 0.458 & 0.426 & 0.516 & 0.510 & 0.445 & 0.510 & 0.413 & 0.445 \\
\texttt{EleutherAI/pythia-12b}  & 1000   & 0.574 & 0.632 & 0.484 & 0.665 & 0.529 & 0.671 & 0.368 & 0.510 \\
\texttt{EleutherAI/pythia-12b}  & 2000   & 0.761 & 0.942 & 0.606 & 0.884 & 0.677 & 0.897 & 0.368 & 0.716 \\
\texttt{EleutherAI/pythia-12b}  & 4000   & 0.858 & 0.994 & 0.684 & 0.916 & 0.774 & 0.942 & 0.355 & 0.806 \\
\texttt{EleutherAI/pythia-12b}  & 8000   & 0.877 & 0.994 & 0.748 & 0.961 & 0.716 & 0.948 & 0.252 & 0.768 \\
\texttt{EleutherAI/pythia-12b}  & 16000  & 0.897 & 0.994 & 0.787 & 0.987 & 0.729 & 0.948 & 0.277 & 0.761 \\
\texttt{EleutherAI/pythia-12b}  & 32000  & 0.929 & 0.994 & 0.755 & 0.987 & 0.723 & 0.968 & 0.290 & 0.761 \\
\texttt{EleutherAI/pythia-12b}  & 64000  & 0.923 & 0.987 & 0.781 & 0.987 & 0.716 & 0.955 & 0.265 & 0.774 \\
\texttt{EleutherAI/pythia-12b}  & 128000 & 0.923 & 1.000 & 0.787 & 0.994 & 0.742 & 0.942 & 0.265 & 0.729 \\
\texttt{EleutherAI/pythia-12b}  & 143000 & 0.923 & 1.000 & 0.774 & 0.981 & 0.735 & 0.948 & 0.290 & 0.735 \\ \bottomrule
\end{tabular}
\caption{Pythia 12B scores on the English-language tasks.}
\label{tab:pythia_12b_score}
\end{table*}

\end{document}